\documentclass{preprint}

\usepackage[T1]{fontenc}
\usepackage{dfadobe}  

\biberVersion
\BibtexOrBiblatex
\usepackage[backend=biber,bibstyle=preprint,citestyle=alphabetic,backref=true, natbib=true, giveninits=true]{biblatex} 
\electronicVersion
\PrintedOrElectronic

\ifpdf \usepackage[pdftex]{graphicx} \pdfcompresslevel=9
\else \usepackage[dvips]{graphicx} \fi

\usepackage[table]{xcolor}
\usepackage{booktabs}
\usepackage[noend,linesnumbered]{algorithm2e}
\RestyleAlgo{ruled}
\SetKwComment{Comment}{/* }{ */}
\usepackage{nicefrac}
\usepackage{amsmath}
\usepackage{amssymb}
\usepackage{mathtools}

\usepackage{colortbl}
\usepackage{pgfplots}
\usepgfplotslibrary{groupplots}
\usepackage{tikz}
\usetikzlibrary{fadings}
\usetikzlibrary{arrows.meta}
\usepackage{makecell}
\usepackage{siunitx}
\usepackage{contour}
\usepackage{multirow}
\usepackage[noend]{algorithmic}
\usepackage{wrapfig}
\usepackage{microtype}
\usepackage{stfloats}

\usepackage[normalem]{ulem}

\definecolor{jorange}{rgb}{8,.5,.0}
\definecolor{bluegreen}{rgb}{.05,.6,.7}
\definecolor{twred}{rgb}{0.85,0,0.2}
\definecolor{twgreen}{rgb}{0,0.65,0.2}
\definecolor{twmagenta}{rgb}{0.7,0,0.4}
\definecolor{mhindigo}{rgb}{0.30,0.20,0.75}
\definecolor{phmaroon}{rgb}{0.50,0.00,0.00}

\definecolor{Crimson}{RGB}{139, 0, 0} %
\definecolor{DarkGoldenrod}{RGB}{218, 165, 32} %
\definecolor{SteelBlue}{RGB}{70, 130, 180} %

\def\clap#1{\hbox to 0pt{\hss #1\hss}}%
\def\initials#1{\protect\clap{\smash{\raisebox{1.4ex}{\tiny{\textsf{\textit{~~#1}}}}}}}%
\makeatletter
\newcommand{\NOTE}[3]{\protect\@ifundefined{hidecomments}{%
  \strut{\color{#2}{\hspace{0pt}\initials{#1}\protect{{\small$\lfloor$}#3{\small]}}}}%
  }{}}
\newcommand{\EDITbyauthor}[4][]{\protect\@ifundefined{hidecomments}{%
  \strut{\color{#3}{\hspace{0pt}\initials{#2}\protect\sout{#1}{#4}}}%
  }{#4}}
\newcommand{\EDITbyauthorred}[4][]{\protect\@ifundefined{hidecomments}{%
  \strut{\color{#3}{\hspace{0pt}\initials{#2}\protect\color{twred}\sout{#1}{\color{#3}{#4}}}}%
  }{#4}}
\newcommand{\EDITredandgreen}[4][]{\protect\@ifundefined{hidecomments}{%
  \strut{\color{twred}{\hspace{0pt}\protect\sout{#1}{\color{twgreen}{#4}}}}%
  }{#4}}
\newcommand{\EDITgreenonly}[4][]{\protect\@ifundefined{hidecomments}{%
  \strut{\color{twgreen}{#4}}%
  }{#4}}

\newcommand{\NOTEboxed}[3]{\protect\@ifundefined{hidecomments}{%
  {\centering\fbox{\parbox{0.97\linewidth}{\protect\NOTE{#1}{#2}{#3}}}}%
  }{}}

\def\myhrulefill{\leavevmode\leaders\hrule height 2pt\hfill\kern\z@}

\makeatother

\def\ignore#1{}

\usepackage{bm}

\newcommand{\fakeparagraph}[1]{\par\vspace{2pt plus 4pt}\noindent{\textbf{#1}}\quad\ignorespaces} %

\renewcommand{\eqref}[1]{Eq.~(\ref{#1})}
\newcommand{\figref}[1]{Fig.~\ref{#1}}
\newcommand{\secref}[1]{Sec.~\ref{#1}}
\newcommand{\tblref}[1]{Tab.~\ref{#1}}

\newcommand{\algoref}[1]{Alg.~\ref{#1}}

\definecolor{bestgreen}{RGB}{0,120,0}
\definecolor{secondgreen}{RGB}{180,230,180}

\definecolor{cropOne}{RGB}{224, 160, 234}
\definecolor{cropTwo}{RGB}{221, 161, 111}
\definecolor{skycyan}{rgb}{0.0, 0.83, 1.0}

\definecolor{occlusion_yellow}{HTML}{FADA24}
\definecolor{occlusion_green}{RGB}{0,170,68}
\definecolor{occlusion_red}{RGB}{199,32,77}
\definecolor{occlusion_purple}{RGB}{202,195,255}
\definecolor{occlusion_blue}{RGB}{14,0,144}

\usetikzlibrary{fit, shapes.geometric}

\newcommand{\spacetargetview}{\mathcal{T}} %
\newcommand{\spaceauxiliary}{\mathcal{A}} %

\newcommand{\sparsedepth}{\mathcal{S}} %
\newcommand{\densedepth}{\mathcal{D}} %

\newcommand{\mask}{\mathcal{M}} %
\newcommand{\occlusionmask}{\mathcal{O}} %
\newcommand{\kernel}{\mathcal{K}} %
\newcommand{\weight}{\mathcal{W}} %
\newcommand{\gaussian}{G_{\sigma}} %

\newcommand{\generalimage}{\mathcal{I}} %
\newcommand{\generalpointcloud}{\mathcal{P}} %
\newcommand{\novelview}{\generalimage^\otimes_\spacetargetview}

\newcommand{\lowband}{\mathcal{L}} %
\newcommand{\highband}{\mathcal{H}} %

\newcommand{\neighborhood}{\mathcal{N}} %

\newcommand{\varcounter}{i} %

\newcommand{\numauxiliary}{k} %
\newcommand{\numsomething}{n} %
\newcommand{\level}{l} %
\newcommand{\timestep}{t} %

\newcommand{\pixelposition}[1][]{{(x_{#1}, y_{#1})}} %

\newcommand{\inspace}[2]{{#1}_{#2}}
\newcommand{\setof}[1]{\{{#1}\}}
\newcommand{\visible}[1]{\tilde{#1}}

\let\titleold\title
\renewcommand{\title}[1]{\titleold{#1}\newcommand{\thetitle}{#1}}
\def\maketitlesupplementary
   {
   \newpage
       \twocolumn[
        \centering
        \Large
        \textbf{\thetitle}\\
        \vspace{0.5em}Supplementary Material \\
        \vspace{1.0em}
       ] %
   }

\title{DIRT: Depth-completed Image-based Rendering using Two-band Blending for Live Novel-view Synthesis}

\author[M.Harrer et al.]
{\parbox{\textwidth}{\centering Mathias Harrer\orcid{0009-0006-7776-4772}, 
        Laura Fink\orcid{0009-0007-8950-1790},
        Noah Lewis\orcid{0009-0000-4446-6994},
        Paul Himmler\orcid{0009-0000-3825-045X}
        and Tim Weyrich\orcid{0000-0002-4322-8844}
        }
        \\
{\parbox{\textwidth}{\centering Friedrich-Alexander-Universität Erlangen-Nürnberg
       }
}
}

\expandafter\ifx\csname iflatexml\endcsname\relax\IfFileExists{latexml.sty}{\usepackage{latexml}}{\expandafter\newif\csname iflatexml\endcsname}\fi%
\begin{document}

\maketitle
\begin{abstract}
State-of-the-art novel-view synthesis (NVS) methods commonly require per-scene optimization before rendering, limiting their applicability in live settings. %
Live or ad-hoc NVS, however, demands low-latency rendering during capture and strong generalization to new scenes. %
Some previous work tried to address this task by adapting training- and optimization-based methods, trading quality for low latency. %
We introduce a live NVS approach following traditional image-based rendering that operates on a posed set of depth sensor observations and images. %
Our method leverages explicit, sparse point-based geometry obtained from active sensors, resulting in a sparsely filled depth buffer that is completed in the (novel) target view's screen-space. %
Using this proxy, a posed set of images is projected into the target frame before fusing the resulting observations through a frequency-aware two-band blending strategy. %
The pipeline is entirely non-neural and optimization-free, requiring no training data or scene-specific optimization. %
As a result, our approach reduces the delay between capture and novel-view rendering and does not rely on favorable network generalization. %
We validate the method's applicability to heterogeneous sensor setups\,---\,including handheld RGB-D cameras, automotive LiDAR, and thermal cameras\,---\,as well as challenging scenarios, such as low light captures. %
Our experiments demonstrate that the approach outperforms prior methods in live reconstruction scenarios, while maintaining competitive performance in conventional NVS benchmarks. %
\end{abstract}
\begin{figure}[t]\iflatexml\includegraphics[width=\linewidth]{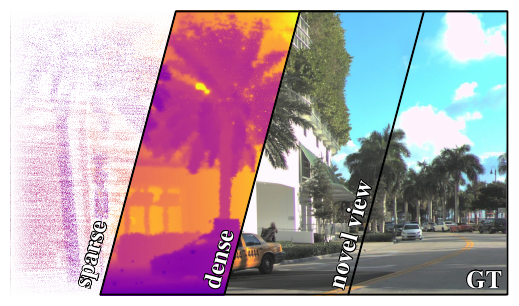}%
\else\par\setbox0\hbox{\kern-5pt\raisebox{-5.0pt}{\includegraphics{figures_tikz/paper_fig0.pdf}}\kern-5pt}\ht0=\dimexpr\ht0-5pt\relax\dp0=\dimexpr\dp0-5pt\relax\box0\vskip-9.0pt\relax\prevdepth=0.0pt\relax\fi%
\end{figure}

\begin{figure*}[!ht]\iflatexml\includegraphics[width=\linewidth]{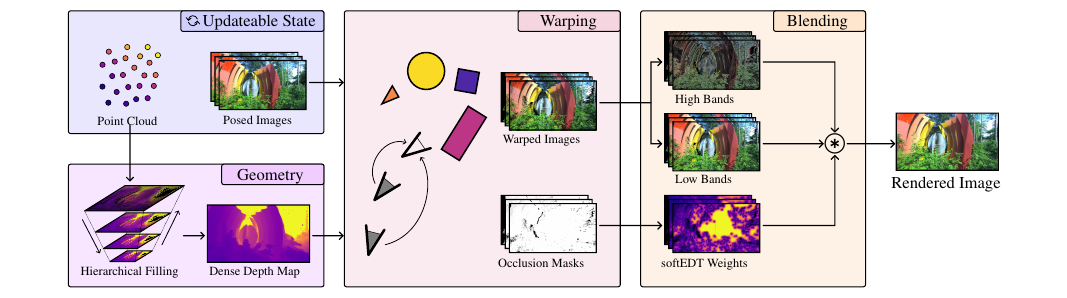}%
\else\par\setbox0\hbox{\kern-5pt\raisebox{-5.0pt}{\includegraphics{figures_tikz/fig_pipeline.pdf}}\kern-5pt}\ht0=\dimexpr\ht0-5pt\relax\dp0=\dimexpr\dp0-5pt\relax\box0\prevdepth=0.0pt\relax\fi%
\caption{\label{fig:pipeline}%
           Pipeline architecture. Given posed images and depth observations, our method produces novel views in milliseconds. %
           First, points are rendered to the target view's screen space to construct a dense depth map through hierarchical depth completion. %
           Auxiliary images are then warped onto the screen-space geometry, while resolving disocclusions. %
           Finally, the images are blended using a two-band approach, weighted by an approximate distance transform as blending weight. %
           }
\end{figure*}

\section{Introduction}
Recent advances in novel-view synthesis (NVS) have demonstrated stunning visual quality, enabling photorealistic rendering of scenes from previously unseen viewpoints~\cite{franke2024trips,barron2022mipnerf360,kerbl20233Dgaussians}.
These approaches typically utilize a dataset comprising multiple images as input to optimize a detailed intermediate scene representation that is subsequently leveraged to render the scene from novel viewpoints.
High visual quality, however, is often achieved at the expense of lengthy scene-specific optimization procedures, significantly extending the time between scene capture and re-rendering.
To overcome this constraint, generalizable neural rendering networks are  explored which amortize per-scene optimization, enabling immediate, high-quality view synthesis~\cite{harrerfranke2023inovis,riegler2021stable, suhail2022generalizablepatch}.
These networks, however, are contingent on favorable generalization to new datasets.
Recently, 3D Gaussian Splatting (3DGS)~\cite{kerbl20233Dgaussians} has been applied in an on-the-fly manner by~\citet{meuleman2025onthefly}:
3D Gaussians are predicted online and optimized concurrently while rendering novel views. %
Given the real-time constraints, their initial rendering starts out commendable, but the ongoing optimization process is still noticeable, and image quality and optimization duration remain tightly linked.
In contrast, our work omits iterative optimization as well as neural components within the NVS pipeline, thereby enabling more efficient resource allocation. %
We target scenarios involving simultaneous acquisition and re-rendering, a process that enables an observer to commence exploration of a scene while capture is still ongoing.
Instead of solely focusing on RGB-only camera setups, we demonstrate how multi-sensor configurations enhance visual fidelity by leveraging data from different modalities. 
Therefore, we specifically target scenarios where multiple cameras, in addition to depth sensors (e.g., LiDAR sensors) are operating simultaneously. 

Our framework maintains an updating scene state that holds a posed set of images and an accumulated point cloud fused from depth sensor observations.
For rendering, the point cloud is projected to a target view and completed into a dense depth map, refining sparse data via a hierarchical filling strategy anchored to real-world measurements. %
By using the resulting dense depth as proxy geometry for multi-view two-band blending~\cite{brown2003recognising}, we facilitate accurate NVS. %

We evaluate our method qualitatively and quantitatively across these challenging settings and via conventional NVS benchmarks, measuring the suitability for live settings by the latency between the arrival of sensor data and view synthesis completion. %
We further illustrate its versatility across diverse datasets: handheld RGB-D captures, night-time driving scenarios, LiDAR-augmented multi-camera rigs, and thermal imagery. Our results show that a fully non-neural image-based rendering (IBR) pipeline remains competitive with state-of-the-art NVS methods amid dominant per-scene optimization and feed-forward learned approaches, attaining superior image quality and faster rendering in \textit{live} settings, and competitive results on established \textit{offline} benchmarks.

In summary, we propose a computation-efficient approach for live-synthesis of high-quality, novel views from sparse spatial data.
Distinctive characteristics and contributions are
\begin{itemize}
    \item the ability to transform sparse point clouds into a high-quality dense depth map through a measurement-anchored hierarchical depth completion;
    \item the seamless integration of established two-band blending in a NVS context;
    \item the inherent architectural flexibility to support heterogeneous sensor inputs and various data modalities without relying on optimization or neural network training.
\end{itemize}

\section{Related Work}
Many NVS methods assume that sufficiently complete observations are available before rendering begins~\cite{fink2023livenvs}.
In live settings, novel views must be synthesized while  acquiring observations, with computational resources shared among tracking, generation of scene representation, and rendering itself.
Accordingly, we review related work by its suitability for online operation, distinguishing optimization-based, scene-agnostic, and sensor-assisted approaches, as well as methods tailored to challenging domains such as driving.

Methods such as NeRFs~\cite{mildenhall2021nerf, barron2021mipnerf, barron2023zip}, 3D Gaussian Splatting~\cite{kerbl20233Dgaussians, hahlbohm2026faster-gs, wu2024recentgs}, or other neural rendering approaches~\cite{ruckert2022adop, aliev2020neural, thies2019deferred, franke2024trips} achieve high-quality NVS by optimizing a scene-specific representation before rendering.
While this is highly effective when observations can be processed offline, the optimization phase often requires minutes to hours, making it poorly suited to live operation.
Several works consequently accelerate per-scene reconstruction in order to produce useful renderings as early as possible~\cite{muller2022instant,meuleman2025onthefly,tosi2026nerfgsslamreview}.
\citet{meuleman2025onthefly} encapsulate feed-forward prediction and optimization of gaussian primitives \textit{on-the-fly}, to enable live NVS.
Despite these advances, accelerated optimization-based methods still require computational effort to build and continually update a scene-specific representation.
In a live system, this computation competes with rendering, camera tracking, and the processing of newly arriving observations. %

This motivates a complementary direction in NVS, rooted in image-based rendering~\cite{shum2008image}, that seeks to synthesize novel views without per-scene optimization. %
Recently, such methods have increasingly been augmented with neural methodologies by incorporating learned components~\cite{hedman2018deep,ha2025geometry,riegler2021stable,rakhimov2022npbg++,harrerfranke2023inovis,fink2023livenvs}, often relying on sensor-based depth observations such as RGB-D or point cloud input.
In foundational work by~\citet{hedman2018deep} blending weights are predicted by a CNN to perform image synthesis.
\citet{ha2025geometry} extend this idea for dynamic scenes captured by multi-view video in controlled studio settings. 
Similar to our approach, they use per-view proxy geometry by fusing a temporal consistent TSDF~\cite{curless1996volumetric} which guides the image blending network. %
\citet{fink2023livenvs} fuse depths from a live RGB-D stream, before projecting neural features into a target-view feature map, subsequently decoding into a novel view by a generalizable CNN.
\citet{harrerfranke2023inovis} do so similarly, but leverage a point-based representation.%

Others adopt target-view-aligned multi-plane images~(MPI)~\cite{mildenhall2019llff,flynn2024quark}, building upon two-plane light fields~\cite{gortler1996lumigraph}.
\citet{mildenhall2019llff} show that MPI parameters can be predicted in a generalizable way.
Further, \citet{flynn2024quark} enable real-time NVS through an iterative render-and-refine architecture with tailored attention operations, at the cost of extensive multi-GPU training.

Recent advances in transformer-based architectures have enabled feed-forward prediction of explicit 3D geometry that can be readily used for NVS.
Several directly regress per-pixel Gaussians~\cite{jiang2025anysplat,zhang2024gslrm,lin2025depthanything3,ziwen2025longlrm,kang2026ilrm,imtiaz2025lvt,zhao2025erayzer,tian2025drivingforward} or surface representations~\cite{wei2024meshlrm,fink2026fus3d}.
A related line of work instead casts image synthesis as a decoding task, rendering views from ray-based queries and encoded image features~\cite{wang2021ibrnet,suhail2022generalizablepatch,jin2024lvsm,szymanowicz2026lagernvs,jiang2025rayzer}.
All of these approaches have substantial training demands and access to large-scale multi-view datasets.

Our method retains explicit geometry to deliver scene-agnostic, high-quality NVS, without relying on learned components. %
Rather than reconstructing geometry from image evidence alone, we exploit data from sensors such as LiDAR or depth cameras. %
This design enables interactive rendering during capture while retaining quality that is competitive with optimization-based and neural methods alike.

Natural applications of NVS under our set of constraints are automotive settings or high-quality previews for handheld RGB-D capturing.
Driving scenarios pose a particularly challenging setting for offline and online NVS~\cite{tian2025drivingforward, zhou2024drivinggaussian, yang2025drivingview, hess2025splatad}. 
Building a sufficiently complete representation requires fusing heterogeneous observations from multiple cameras and, often, additional sensors such as LiDAR~\cite{caesar2020nuscenes, sun2020waymo, liao2022kitti}. %
Moreover, data are acquired in uncontrolled outdoor environments containing occlusions, moving objects, and substantial appearance variation~\cite{zhou2025hugsim, sun2020waymo, caesar2020nuscenes}.
Finally, the capture trajectory is dictated by vehicle motion and the road network, resulting in predominantly street-level viewpoints and incomplete coverage of the scene~\cite{caesar2020nuscenes, sun2020waymo, wilson2021argoverse2}, making view extrapolation specifically challenging~\cite{zhou2025hugsim, yang2025drivingview}.
Due to these challenges, existing methods get specifically tailored to automotive settings, hindering their use in broader NVS scenarios.
In case of handheld capture systems, live previews are central, communicating coverage, progress, and potential failures to users~\cite{fink2023livenvs, meuleman2025onthefly}. However, for capture guidance, visually plausible learned synthesis~\cite{harrerfranke2023inovis, fink2023livenvs} is not necessarily desirable: prior-driven predictions may hide missing observations or acquisition artifacts. 
Conversely, our approach offers a visually appealing, yet deterministic rendering that unveils the current state of capture.

\section{Depth-completed Image-based Rendering}
\label{sec:method}
Inspired by preceding IBR pipelines~\cite{buehler2001unstructured, debevec1998efficient}, our approach comprises three main steps to perform view synthesis: %
initially, we construct a target view aligned geometric depth proxy (\secref{sec:geometry}). 
This proxy enables projection of auxiliary images to the target view during the subsequent warping step (\secref{sec:warping}). %
Finally, data from multiple, projected auxiliary images is blended to create a novel view (\secref{sec:blending}).
An overview of our method is presented in \figref{fig:pipeline}.

Our method assumes posed images~$\setof{\generalimage}$ combined with an unstructured point cloud~$\generalpointcloud$ as input. %
We denote the target space as $\spacetargetview$ and the auxiliary images as $\setof{\inspace{\generalimage}{\spaceauxiliary}}^\numauxiliary_{\varcounter=1...\numauxiliary} \subseteq \setof{\generalimage}$. 
While our approach allows the number of auxiliary images, $\numauxiliary$, to be freely chosen, we generally set $\numauxiliary=4$ as practical trade-off between coverage and overall performance, see supp. material~(S5).
Note that the approach neither assumes $\setof{\generalimage}$ are captured by a single camera or modality, nor do they require uniform aspect ratio or resolution. %
\subsection{Geometric Proxy}
\label{sec:geometry}
\fakeparagraph{Sparse Depth Reconstruction.}
The input point cloud~$\generalpointcloud$ can be provided as a complete cloud for the entire scene, or accumulated from a continuous stream of consecutive depth observations, like single depth images or LiDAR sweeps. %
In the latter case, our accumulation is inspired by \citet{keller2013Realtime}, filtering out transient or dynamic objects.
We extend \citet{keller2013Realtime} by a spherical camera model for LiDAR sensors and condense the point cloud structure, omitting normals and radii. %
When rendering a target view~$\inspace{\generalimage}{\spacetargetview}$, this accumulated point cloud~$\generalpointcloud$ is combined with the most recent partial observation to form $\generalpointcloud^\timestep$,
allowing to roughly approximate scene dynamics.
$\generalpointcloud^\timestep$ is rendered to the target frame's image space.
This yields a sparse 2D depth map $\sparsedepth$ of the target frame with unresolved point visibility. 
In order to increase coverage and avoid artifacts in background regions, additional points are generated within these areas at a maximum depth value.
We resolve visibility based on $\generalpointcloud^\timestep$'s projection in 2D image space, as it focuses processing on the potentially visible point set rather than costly computation over the whole, large-scale point cloud in 3D.
\begin{algorithm}[t]
    \linespread{0.95}\selectfont
    \footnotesize
    \caption{\textsc{completeDepth}($\sparsedepth$)\label{alg:depthcompletion}}
    \KwData{$\sparsedepth^\level:$ (sparse) depth map, size $w \times h$ .}
    \KwResult{$\densedepth^\level:$ dense depth map}
\begin{algorithmic}[1]
    \IF{$w \geq 2 \land h \geq 2$}
    
        \DontPrintSemicolon
    
        \STATE $ \mask \gets 1 \textbf{ where } \sparsedepth^\level \text{ valid } \textbf{else } 0$ \\
        \STATE $\neighborhood \gets \frac{\kernel \ast \sparsedepth^\level}{\kernel \ast \mask}$ \\
        \STATE \nllabel{alg:depthcompletion:filtering}
            $\mask \gets \neg \mask \lor \sparsedepth^\level > \neighborhood\cdot\tau^\level$\\
        \STATE $\sparsedepth^{\level+1} \gets \text{downsample}(\sparsedepth^\level \cdot \neg \mask)$\\
        \STATE $\densedepth^{\level+1} \gets \textsc{completeDepth}(\sparsedepth^{\level+1})$ \nllabel{alg:depthcompletion:downsampling} \\
        \STATE $\densedepth^{\level+1}  \gets \text{bilateral}(\densedepth^{\level+1})$ \nllabel{alg:depthcompletion:filling}\\
        \STATE $\densedepth^\level \gets \text{upsample}(\densedepth^{\level+1})$ \\
        \STATE $\densedepth^\level \gets \sparsedepth^\level \cdot \neg \mask + \densedepth^\level \cdot \mask $
    \ELSE
        \STATE $\densedepth^\level \gets \sparsedepth^\level$
    \ENDIF
    \RETURN $\densedepth^\level$
    \end{algorithmic}

\end{algorithm}
\fakeparagraph{Depth Completion.}
In order to resolve holes and faulty point visibility present in $\sparsedepth$, we apply a recursive image-space depth completion inspired by \citet{grossman1998point}.
Unlike learned depth completion methods, which densify sparse measurements via models trained on domain-specific benchmarks~\cite{uhrig2017sparsity, park2020nonlocal} and may hallucinate geometry where observations are absent, our completion remains classical and measurement-anchored, requiring no training and transferring across sensors unadapted.
To construct the dense depth map $\densedepth$, we face two main challenges: 
first,  missing data has to be predicted based on the surrounding samples; and second, determining whether depth samples belong to visible geometry or should be occluded. %
\algoref{alg:depthcompletion} outlines our approach to address these challenges.  The two following paragraphs detail its subcomponents.

\fakeparagraph{Depth Completion:\;Hierarchical Filling.}
To predict missing data from the available samples, we repeatedly filter occluded points and downsample $\sparsedepth$ , whereby each downsampled depth map level $\sparsedepth^\level$ will contain less holes than the previous. %
To do so, we first filter out occluded points (next paragraph) and then downsample by averaging valid samples in a $2 \times 2$ neighborhood.
Thus, we acquire the next lower resolution of the depth map $\sparsedepth^{\level+1}$ (\algoref{alg:depthcompletion}:\ref{alg:depthcompletion:downsampling}).
We repeat the occlusion filtering and downsampling steps interleaved until we reach the highest possible level~$\level$.
This extracts low frequency but dense information from the sparse input data. %
The resulting depth map hierarchy is fused back together by progressively upsampling, executing the following procedure at every level: %
firstly, a bilateral smoothing is applied to $\sparsedepth^{\level+1}$ yielding $\densedepth^{\level+1}$, which prevents staircasing, following work by~\citet{jimenez2014next}; %
secondly,  the image is bilinearly upsampled to acquire a dense depth map $\densedepth^{\level}$;
and thirdly, $\densedepth^{\level}$ is fused with high frequency detail from $\sparsedepth^{\level}$.
Thus, low-frequency information from coarser resolution levels fills missing data in higher resolutions~(\algoref{alg:depthcompletion}:\ref{alg:depthcompletion:filling}). %
\fakeparagraph{Depth Completion:\;Occluded Point Filtering.}
Due to the sparsity of point clouds, the depth map~$\sparsedepth^\level$ on any level may display points of occluded scene content.
We filter such occluded samples by estimating the depth of each pixels local neighborhood~$\neighborhood^\level$ and discard samples that are significantly more distant.
This is done by applying a zero-centered high-pass kernel $\kernel$ (see supp. material~(S1.1) for details) to $\sparsedepth^\level$ at each level $\level$ (\algoref{alg:depthcompletion}:\ref{alg:depthcompletion:filtering}).
For brevity, we omit the level in the following descriptions of the filtering step whenever feasible.
The neighborhood~$\neighborhood$ is given by:
\begin{align}
    \small%
    \neighborhood = 
    \frac{\kernel \ast \sparsedepth}{\kernel \ast \mask}, 
    &&\text{with } \mask\pixelposition = \begin{cases}
        1 & \text{if } \sparsedepth\pixelposition \text{ is set}\\
        0 & \text{else} \quad\quad\quad\;,\\
    \end{cases}\nonumber\\[-1.0\baselineskip]
\end{align}
where $\mask$ represents the mask of valid points.
The result of the high-pass filtering is normalized using $\mask$ to ensure that only values contribute where the corresponding pixel $\pixelposition$ in $\sparsedepth$ has a valid depth value. 
This effectively measures the local neighborhoods depth, thereby allowing for filtering of points that demonstrate a high degree of deviation.
If the neighborhood $\neighborhood$ is found to be significantly closer to the camera than the current point, the point is considered to be occluded and gets discarded by setting its value to 0.
\begin{align}
    \small
    \sparsedepth\pixelposition = \begin{cases}
        \sparsedepth\pixelposition & \text{if } \sparsedepth\pixelposition \leq \neighborhood\pixelposition\cdot\tau^\level\\
        0 & \text{else} \quad\quad\quad\quad\quad\quad\quad\quad\quad\quad.\\
    \end{cases}
\end{align}
The parameter $\tau^\level\geq1$ controls the sensitivity for sample rejection.
(In our experiments, we found $\tau^\level = 1 + 5^\level\cdot0.02$ strikes a good balance between the preservation of detail and the avoidance of occlusion based geometric inaccuracies.)
Note that while the methodology is mathematically defined in linear depth parametrization for clarity, the implementation uses inverse depth for correct interpolation. 

\subsection{Warping and Disocclusion Detection}
\label{sec:warping}
\begin{figure}[t]\iflatexml\includegraphics[width=\linewidth]{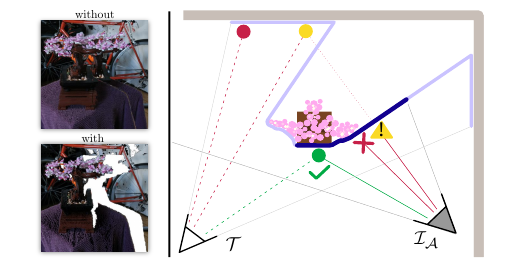}%
\else\par\setbox0\hbox{\kern-5pt\raisebox{-5.0pt}{\includegraphics{figures_tikz/fig_disocclusion.pdf}}\kern-5pt}\ht0=\dimexpr\ht0-5pt\relax\dp0=\dimexpr\dp0-5pt\relax\box0\prevdepth=0.0pt\relax\fi%
\caption{\label{fig:disocclusion}%
           Depth test to resolve disocclusions. 
            When fetching color from auxiliary views $\inspace{\generalimage}{\spaceauxiliary}$ to target view $\spacetargetview$,
            depth map $\densedepth$ is rendered as a mesh to $\inspace{\generalimage}{\spaceauxiliary}$ (\textcolor{occlusion_purple!80!black}{light purple}: $\densedepth$ as mesh; \textcolor{occlusion_blue}{blue}: part of mesh visible in $\inspace{\generalimage}{\spaceauxiliary}$): \textcolor{occlusion_green}{green}: test passes; \textcolor{occlusion_red}{red}: test fails $\rightarrow$ occlusion; \textcolor{occlusion_yellow}{yellow}: false negative $\rightarrow$ test fails but point should be visible.   
           }
\end{figure}
After constructing the geometric proxy in the form of the dense depth map $\densedepth$, a subset of $\numauxiliary$ auxiliary images $\setof{\inspace{\generalimage}{\spaceauxiliary}}$ is projected onto the recovered geometry. %
Conventionally, this is achieved through one of two methods. %
In a forward approach, auxiliary pixels are warped to the proxy geometry, effectively splatting information
to the target view $\spacetargetview$. %
Hence, depth data for each auxiliary image is required. %
Conversely, in a backward approach, target pixels are warped to auxiliary images, from which information is retrieved. %
Given that acquiring depth data for each view entails a high computational cost, the latter is better suited to real-time constraints. %
However, backward warping is susceptible to disocclusion artifacts, whereby target pixels may be mapped to areas occluded in the auxiliary view $\inspace{\generalimage}{\spaceauxiliary}$. %
Therefore, these pixels require explicit handling to prevent misprojections.
We unproject the depth map~$\densedepth$ for each pixel position~$\pixelposition$, yielding a point set~$p \in \inspace{\generalpointcloud}{\spacetargetview}$ .
This point set is used in two ways:
we first construct a continuous scaffold in the form of a triangle mesh from $\inspace{\generalpointcloud}{\spacetargetview}$ by connecting points of neighboring pixels to form triangles.
By rendering this scaffold from the perspective of each $\inspace{\generalimage}{\spaceauxiliary}$, we get a new approximate dense depth map~$\inspace{\densedepth}{\spaceauxiliary}$.
Then, we use the per-pixel points~$\inspace{\generalpointcloud}{\spacetargetview}$ a second time, projecting them to the auxiliary views as well.
This yields the depths $d_p$ of the projected points in the perspective of each auxiliary image~$\inspace{\generalimage}{\spaceauxiliary}$.
The subset $\visible{\inspace{\generalpointcloud}{\spaceauxiliary}}$, visible in $\inspace{\generalimage}{\spaceauxiliary}$, is then determined by
{\small
\begin{align}
    \visible{\inspace{\generalpointcloud}{\spaceauxiliary}} = \{p \in \inspace{\generalpointcloud}{\spacetargetview}: d_p - \inspace{\densedepth}{\spaceauxiliary}\pixelposition[p] < \epsilon \}.
\end{align}}%
$\pixelposition[p]$ are $p$'s pixel positions in $\inspace{\generalimage}{\spaceauxiliary}$, and $\epsilon$ is a depth and normal dependent interval to account for numerical errors.
Finally, we get the warped image $\inspace{\generalimage}{\spacetargetview}$ in the target view space by fetching the color of $\inspace{\generalimage}{\spaceauxiliary}$ at $\pixelposition[\visible{p}]$ of $\visible{p} \in \visible{\inspace{\generalpointcloud}{\spaceauxiliary}}$ and write it to the pixel position~$\pixelposition$.
Because colors are only fetched for pixels that belong to  visible points, disocclusions show as holes in the warped image 
(\figref{fig:disocclusion}).
We track the position of these holes to get the occlusion mask $\inspace{\occlusionmask}{\spacetargetview}$, yielding a set of $\numauxiliary$ warped images $\setof{\inspace{\generalimage}{\spacetargetview}}$ and occlusion masks $\setof{\inspace{\occlusionmask}{\spacetargetview}}$.
Disocclusion handling and depth test are illustrated in~\figref{fig:disocclusion}.

\subsection{Blending}
\label{sec:blending}
The final step in the pipeline is the blending of the warped auxiliary images $\setof{\inspace{\generalimage}{\spacetargetview}}$ to create the novel view.

\begin{figure}[t]\iflatexml\includegraphics[width=\linewidth]{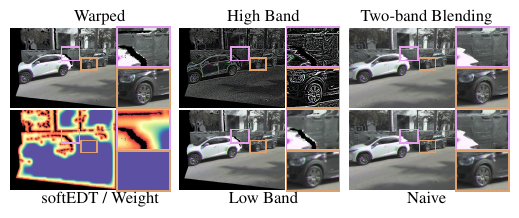}%
\else\par\setbox0\hbox{\kern-5pt\raisebox{-6.32793pt}{\includegraphics{figures_tikz/fig_twoband.pdf}}\kern-5pt}\ht0=\dimexpr\ht0-5pt\relax\dp0=\dimexpr\dp0-5pt\relax\box0\prevdepth=1.32793pt\relax\fi%
\caption{\label{fig:twoband}%
           Naive blending of multiple images causes ghosting and loss of detail. Two-band blending~\cite{brown2003recognising} smooths low frequencies while retaining high-frequency detail, mitigating said artifacts. The high-band image has been adjusted for visibility.}
\end{figure}

\fakeparagraph{Blending Weights.}
Following conventional IBR approaches~\cite{buehler2001unstructured,debevec1998efficient}, a weight image $\weight$ is assigned to each candidate~$\inspace{\generalimage}{\spacetargetview}$.
Our weights build upon the occlusion mask $\inspace{\occlusionmask}{\spacetargetview}$ exclusively.
This occlusion aspect is prioritized, as depth-completed geometry and scene content may be misaligned, resulting in foreground image content projected onto background geometry, or vice versa, particularly at border regions, visible as seams.
As these artifacts occur in the neighborhood of outlines of occlusion masks, they are hereafter denoted as \textit{occlusion edges}.
To minimize these, we employ an approximate, smoothed Euclidean distance transform (\textit{softEDT}; see supp. material~(S1.2)) to estimate a distance function $D$ to the occlusion edges.
$D$ stays faithful near the foreground and smooths increasingly with distance, eliminating the ridges and banding characteristics of a Euclidean distance transform.
The resultant value is then used as blending weights $\weight$, see~\figref{fig:twoband}.
Consequently, pixels near occlusion edges are assigned a reduced weight, while more distant ones are assigned a higher weight. This process facilitates the elimination of seam artifacts while focusing on regions where the information is likely to be reliable.
\fakeparagraph{Two-band Blending.}
Blending images naively with per-pixel weights $\setof{\weight}$, may compromise high frequency detail and result in ghosting.
Such artifacts can be attributed to imperfect pose calibration or deviation of the synthesized geometry from ground truth. Our proposed blending strategy draws on related work by~\citet{brown2003recognising}. Based on foundational work by~\citet{burt1983multiresolution}, they develop a two-band blending approach for stitching panoramas from multiple images mitigating the loss of high-frequencies. This blending strategy is well suited to the challenges inherent in our scenario, including varying aperture or exposure time and vignetting.
The idea is the seamless blending of low-frequency information of all images within a broader spatial area, simultaneously focusing on blending high-frequency detail only in a small spatial neighborhood.
The separation in two frequency bands is achieved through the use of a Laplacian Pyramid~\cite{burt1983laplacian}, leading to fast frequency-aware blending. %
 We propose the application of two-band blending to image-based rendering.
The low-frequency band $\lowband$ for a frame $\inspace{\generalimage}{\spacetargetview}$ is obtained by applying a low-pass filter using a gaussian function $\gaussian$, where $\sigma = 0.95$.
The high-frequency detail for each frame is recovered by subtracting this low-frequency band from the warped input image, yielding
{\small
\begin{align}\label{eq:twoband}
\lowband= \gaussian \ast \inspace{\generalimage}{\spacetargetview} && \highband = \inspace{\generalimage}{\spacetargetview} - \lowband \;.
\end{align}}%
The synthesis of the final image $\novelview$ is then achieved per pixel, whereby the set of low frequencies $\setof{\lowband}$ are blended with the corresponding weights $\setof{\weight}$. The high frequencies are selected from the candidate view of maximum weight.
{\small
\begin{equation}\label{eq:blending}
    \novelview\pixelposition = \highband_{\weight_\text{max}}\pixelposition + \sum_\varcounter^\numauxiliary \setof{\lowband}_\varcounter\pixelposition * \setof{\weight}_\varcounter\pixelposition\;.
\end{equation}}%
As demonstrated in~\figref{fig:twoband}, low frequencies are seamlessly blended, thereby reducing the visibility of image borders, while high-frequency detail is retained.
Finally, non-covered image regions are filled with a simple \textit{PushPull} \cite{gortler1996lumigraph} algorithm, coloring such regions by interpolating nearby data.

\newcommand{\thinline}{\specialrule{0.1pt}{0pt}{0pt}}
\begin{table*}[t]
\footnotesize
\renewcommand{\arraystretch}{1.0}
\setlength{\tabcolsep}{5pt}
\centering
\begin{tabular}{l|ccc|ccc|ccc|ccc}
\toprule
 & \multicolumn{3}{c|}{LiveNVS} & \multicolumn{3}{c|}{nuScenes} & \multicolumn{3}{c|}{Waymo} & \multicolumn{3}{c}{Argoverse 2} \\
 & \multicolumn{3}{c|}{(\scriptsize $1280\times 720$ / $1920\times 1080$)} & \multicolumn{3}{c|}{(\scriptsize $1600\times 900$)}& \multicolumn{3}{c|}{(\scriptsize $1920\times 1280$)} & \multicolumn{3}{c}{(\scriptsize $1800\times 1550$)}\\
\midrule
Approach & PSNR$\uparrow$ & LPIPS$\downarrow$ & Time\,{\tiny \raisebox{0.3ex}{[\si{\milli\second}]}} & PSNR$\uparrow$ & LPIPS$\downarrow$ & Time\,{\tiny \raisebox{0.3ex}{[\si{\milli\second}]}} & PSNR$\uparrow$ & LPIPS$\downarrow$ & Time\,{\tiny \raisebox{0.3ex}{[\si{\milli\second}]}} & PSNR$\uparrow$ & LPIPS$\downarrow$ & Time\,{\tiny \raisebox{0.3ex}{[\si{\milli\second}]}}\\
\midrule
Depth Anything 3 & 15.51 & 0.4909 & 6222 & 17.46 & 0.5638 & 6258 & 19.23 & 0.5343 & 4958 & 16.63 & 0.5135 & 3704\\
Long-LRM & \cellcolor{secondgreen}19.93 & \cellcolor{secondgreen}0.2349 & 2686 & 23.55 & 0.3572 & 2638 & \cellcolor{secondgreen}25.23 & \cellcolor{secondgreen}0.3559 & 1939 & \cellcolor{secondgreen}23.54 & \cellcolor{secondgreen}0.3246 & 2791 \\
DrivingForward & - & -  & - & \cellcolor{bestgreen}\textcolor{white}{24.00} & \cellcolor{secondgreen}0.3221 & 590 & 24.36 & 0.4095 & 284 & 20.21 & 0.5106  &  \cellcolor{secondgreen}285 \\ %
\thinline
Inovis & 19.01 & 0.2736 & 157 & 22.20 & 0.4396 & \cellcolor{secondgreen}146 & 22.49 & 0.4157 & \cellcolor{secondgreen}243 & 19.80 & 0.3763 & 289 \\
LiveNVS & 16.60 & 0.4495  & \cellcolor{bestgreen}\textcolor{white}{49} & - & - & - & - & - & - & - & - & - \\ %
\thinline
On-the-fly NVS & 18.86 & 0.3798 & 639 & \cellcolor{secondgreen}23.56 & 0.4622 & 439 & \cellcolor{bestgreen}\textcolor{white}{25.68} & 0.4270 & 693 & 18.19 & 0.5474 & 486 \\
\thinline
Ours & \cellcolor{bestgreen}\textcolor{white}{22.18} & \cellcolor{bestgreen}\textcolor{white}{0.1959} & \cellcolor{secondgreen}69 & 23.47 & \cellcolor{bestgreen}\textcolor{white}{0.2711} & \cellcolor{bestgreen}\textcolor{white}{71} & 24.79 & \cellcolor{bestgreen}\textcolor{white}{0.2506} & \cellcolor{bestgreen}\textcolor{white}{98} & \cellcolor{bestgreen}\textcolor{white}{25.05} & \cellcolor{bestgreen}\textcolor{white}{0.1802} & \cellcolor{bestgreen}\textcolor{white}{135}\\
\bottomrule
\end{tabular}
\caption{\label{tbl:results_live}Quantitative results for live NVS. A dash indicates that a method was not applicable to a dataset.
}
\end{table*}
\section{Results}

\subsection{Experimental Setup}
All methods are evaluated on the same machine (Intel\,i9-12900\,CPU,\,NVIDIA\,A5000\,GPU\,24\,GB\,VRAM,\,64\,GB\,RAM).
To evaluate image quality, we report PSNR and LPIPS~\cite{zhang2018unreasonable} (with AlexNet~\cite{krizhevsky2012imagenet} backbone).
Given our focus on live NVS, we define the primary performance metric \textit{capture-to-image} as the interval between the arrival of the last input view's sensor data and the subsequent synthesis of the test view.
We benchmark competing methods as faithful as possible.
All evaluations are conducted with the previously described set of hyperparameters, e.g., $\tau^\level$, $\sigma$, and $\numauxiliary$.

\fakeparagraph{Baselines.}
Unless stated otherwise, we evaluate generalizable approaches using their released checkpoints without fine-tuning~\cite{harrerfranke2023inovis,fink2023livenvs,lin2025depthanything3,ziwen2025longlrm,tian2025drivingforward}.
Further details on the setup of baselines is provided in the supp. material~(S2).

We identify \textit{On-the-fly NVS}~\cite{meuleman2025onthefly}, which targets live NVS from unposed RGB streams, as our closest baseline with publicly available code.
To isolate NVS performance, we replace its pose-estimation with ground-truth poses.
We further evaluate the neural IBR methods \textit{LiveNVS}~\cite{fink2023livenvs} and \textit{Inovis}~\cite{harrerfranke2023inovis}.
We compare to \textit{LiveNVS} exclusively on RGB-D data as it requires dense depth maps.
As general feed-forward Gaussian reconstruction baselines, we include \textit{Depth Anything 3}~\cite{lin2025depthanything3} and \textit{Long-LRM}~\cite{ziwen2025longlrm},
as well as \textit{DrivingForward}~\cite{tian2025drivingforward}, which is specifically tailored to driving scenarios.
For offline evaluation, we remove runtime constraints and allow \textit{On-the-fly NVS}~\cite{meuleman2025onthefly} to process all images before re-rendering.
We additionally compare with \textit{Faster-GS}~\cite{hahlbohm2026faster-gs}, which accelerates reconstruction while mirroring the quality of \textit{3DGS}~\cite{kerbl20233Dgaussians}, also with default optimization time.
\fakeparagraph{Datasets.}
For live evaluation, we simulate an evolving capture process using timestamped observations.
Adopting the protocol of~\citet{tian2025drivingforward}, methods receive only observations within the preceding temporal window to render a test view, evaluating extrapolation to the next viewpoint rather than interpolation between observed views.
We select automotive scenes from \textit{nuScenes}~\cite{caesar2020nuscenes}, \textit{Waymo}~\cite{sun2020waymo}, and \textit{Argoverse 2}~\cite{wilson2021argoverse2} with limited dynamic content, as all evaluated methods assume predominantly static scenarios.
For each test frame, methods receive observations from the preceding \(5\) seconds.
While all datasets feature multi-camera rigs, we evaluate on front cameras only, as some competing methods do not support multiple camera streams \cite{harrerfranke2023inovis,meuleman2025onthefly,fink2023livenvs}.
In contrast, \textit{DrivingForward} receives all camera streams, as it is designed for such scenarios.
We evaluate handheld RGB-D capture on the keyframed \textit{LiveNVS} dataset~\cite{fink2023livenvs}, with sequences at roughly one frame per second; methods receive data from the preceding \(15\) seconds to render a test view.

For conventional offline NVS, we evaluate 
on seven \textit{MipNeRF 360}~\cite{barron2022mipnerf360} and four \textit{Tanks and Temples} scenes~\cite{Knapitsch2017}.
To assess cross-modal robustness, we further evaluate on \mbox{thermal} imagery from \textit{RGBT-Scenes}~\cite{lu2024thermalgaussian}.
Extensive details on~data\-set preprocessing can be found in the supp. material~(S3).
\subsection{Evaluation}
\begin{figure*}[p]\iflatexml\includegraphics[width=\linewidth]{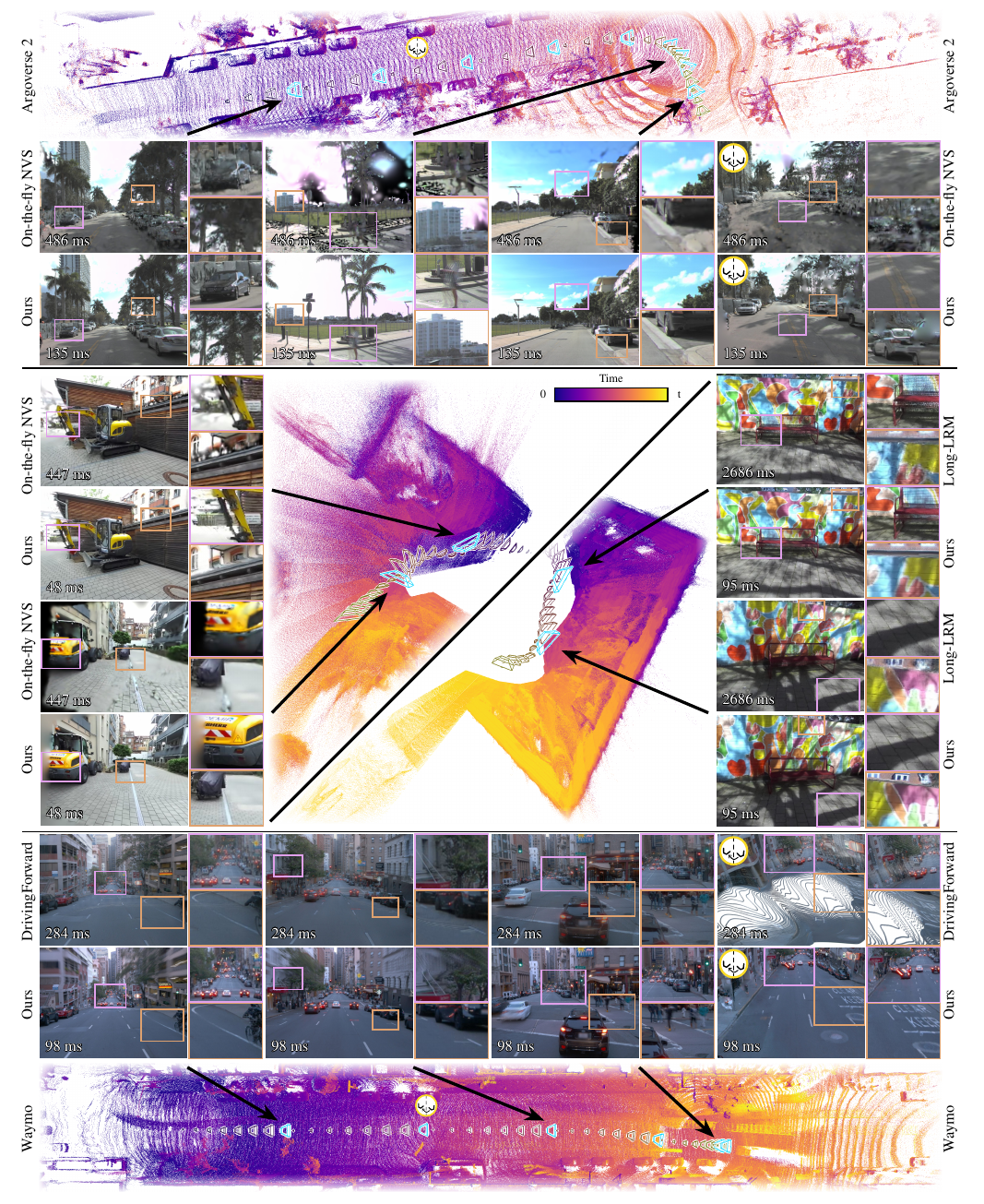}%
\else\par\setbox0\hbox{\kern-5pt\raisebox{-311.98792pt}{\includegraphics{figures_tikz/fig_results_live.pdf}}\kern-5pt}\ht0=\dimexpr\ht0-5pt\relax\dp0=\dimexpr\dp0-5pt\relax\box0\vskip-5.5pt\relax\prevdepth=306.98792pt\relax\fi%
\caption{Qualitative comparison with \textit{On-the-fly NVS}~\cite{meuleman2025onthefly}, \textit{Long-LRM}~\cite{ziwen2025longlrm}  and \textit{DrivingForward}~\cite{tian2025drivingforward}. We state measured \textit{capture-to-image} times. Views taken from trajectories are highlighted by arrows. Extrapolating lane shifts are marked with \includegraphics[height=0.75\baselineskip]{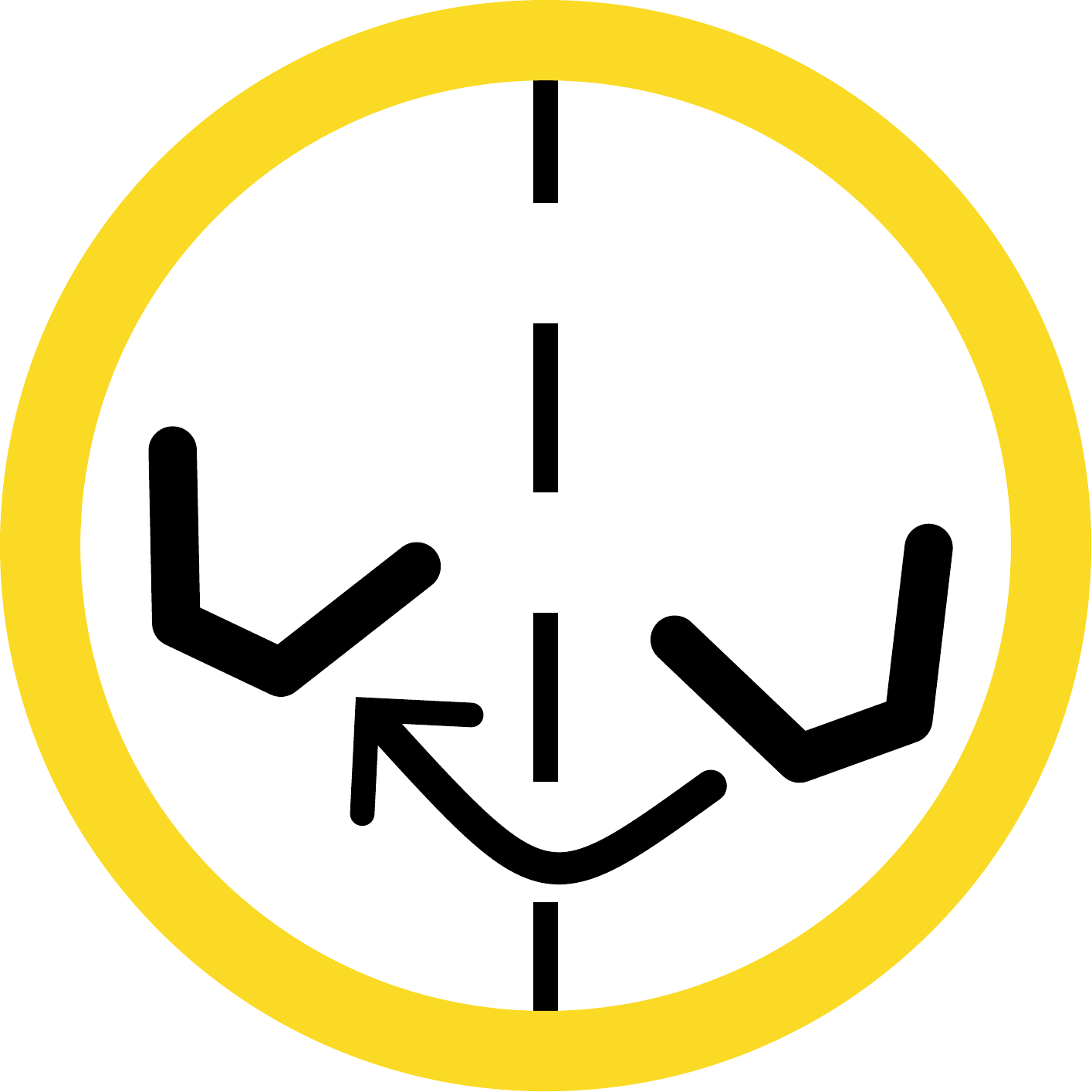}.}%
    \label{fig:results_live}
\end{figure*}

\fakeparagraph{Live.}
Quantitative live-reconstruction results are reported in \tblref{tbl:results_live}.
Our method achieves the best LPIPS across all datasets, competitive PSNR, and substantially lower \textit{capture-to-image} time than baselines with comparable fidelity.
Although \textit{Long-LRM}, \textit{DrivingForward}, and \textit{On-the-fly NVS} achieve higher PSNR on individual nuScenes and Waymo evaluations, ours obtains lower LPIPS in all comparisons.
\figref{fig:results_live} shows representative live automotive and handheld RGB-D results.
Our method produces sharper textured details than \textit{On-the-fly NVS}. %
Though ours can exhibit local artifacts at depth discontinuities, the latter's time-constrained optimization may fail to converge, particularly for small capture baselines.
\textit{Long-LRM} suffers from blurry textures, missing high-frequency detail.
Compared to \textit{DrivingForward}, our method recovers finer detail but can smear in regions poorly covered by depth observations, while they better preserve dynamic objects.
Under a \textit{lane shift} setting, our method yields substantially more plausible views than both \textit{On-the-fly NVS} and \textit{DrivingForward}.

\fakeparagraph{Offline.}
\begin{figure*}[t]\iflatexml\includegraphics[width=\linewidth]{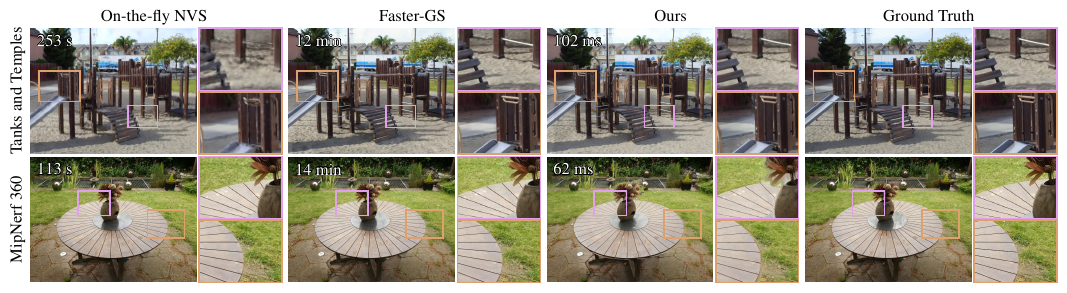}%
\else\par\setbox0\hbox{\kern-5pt\raisebox{-32.93756pt}{\includegraphics{figures_tikz/fig_results_offline.pdf}}\kern-5pt}\ht0=\dimexpr\ht0-5pt\relax\dp0=\dimexpr\dp0-5pt\relax\box0\prevdepth=27.93756pt\relax\fi%
\caption{Qualitative comparison in an offline setting. Total reconstruction time is reported for each method. On-the-fly NVS and Faster-GS processed the scenes completely before rendering, while our approach reconstructs high-quality novel views without training or optimization.}%
    \label{fig:results_offline}
\end{figure*}%
Qualitative and quantitative offline results are shown in \figref{fig:results_offline} and \tblref{tbl:results_offline}.
Our method is competitive with \textit{Faster-GS} and consistently outperforms \textit{On-the-fly NVS}.
Runtime includes full-scene optimization time before rendering.
\textit{On-the-fly NVS} struggles with sharp geometry and planar regions, while ours mainly exhibits seam-like artifacts near depth discontinuities.
This demonstrates our method's flexibility beyond the primary scenario of live NVS. %
Further live and offline qualitative results, plus discussion of the differing metric trends, are provided in the supp. material~(S4).
\fakeparagraph{Multimodal and Challenging Scenarios.}
\figref{fig:multimodal} evaluates night-time automotive scenes and thermal imagery from \textit{RGBT-Scenes}, while \tblref{tbl:results_offline} reports corresponding thermal metrics.
Our method remains robust under low-light conditions, where the optimization-based baselines often fail to converge to a consistent scene representation.
On thermal data, 
our method outperforms \textit{Faster-GS} and \textit{On-the-fly NVS}, demonstrating robustness across sensing modalities.
\fakeparagraph{Ablation:\;Depth Source.}
To emulate RGB-D data, we compare different sources of depth (monocular~\cite{lin2025depthanything3}, transformer-based multi-view stereo (MVS)~\cite{cao2024mvsformer++}) against captured RGB-D stereo~\cite{stereolabs2023depthsettings}. We also evaluate our depth processing against TSDF~\cite{curless1996volumetric,millane2024nvblox} ablating our methods' effectiveness to acquire target view depth from input data (\tblref{tbl:ablation_geometry}).
See supp. material~(S5) for more details, a qualitative comparison, and further ablations on number of input views, point cloud density, depth completion, and blending. %
\begin{table}[t]
\scriptsize
\renewcommand{\arraystretch}{1.0}
\setlength{\tabcolsep}{4pt}
\centering
\begin{tabular}{l|cc|cc|cc}
\toprule
 & \multicolumn{2}{c|}{MipNerf 360} & \multicolumn{2}{c|}{RGBT-Scenes} & \multicolumn{2}{c}{Tanks and Temples} \\
\midrule
Approach & PSNR$\uparrow$ & LPIPS$\downarrow$ & PSNR$\uparrow$ & LPIPS$\downarrow$ & PSNR$\uparrow$ & LPIPS$\downarrow$ \\
\midrule
Faster-GS & \cellcolor{bestgreen}\textcolor{white}{27.31} & \cellcolor{bestgreen}\textcolor{white}{0.1420} & \cellcolor{secondgreen}24.36 & \cellcolor{secondgreen}0.1422 & \cellcolor{bestgreen}\textcolor{white}{22.38} & \cellcolor{secondgreen}0.2449 \\
On-the-fly NVS & 24.22 & 0.2659 & 21.23 & 0.2248 & 19.88 & 0.3928 \\
Ours & \cellcolor{secondgreen}25.20 & \cellcolor{secondgreen}0.1486 & \cellcolor{bestgreen}\textcolor{white}{24.37} & \cellcolor{bestgreen}\textcolor{white}{0.1371} & \cellcolor{secondgreen}22.10 & \cellcolor{bestgreen}\textcolor{white}{0.2365} \\
\bottomrule
\end{tabular}
\caption{\label{tbl:results_offline}Quantitative results for offline NVS on MipNerf 360, Tanks and Temples, and thermal RGBT-Scenes datasets.}
\end{table}
\begin{table}[t]
\scriptsize
\renewcommand{\arraystretch}{1.0}
\setlength{\tabcolsep}{4pt}
\centering
\begin{tabular}{l|c||c|c}
\toprule
Depth Source  & Time per  & TSDF  & Ours  \\
  & Image\,{\tiny \raisebox{0.3ex}{[\si{\milli\second}]}} & LPIPS$\downarrow$\;/\;Time\,{\tiny \raisebox{0.3ex}{[\si{\milli\second}]}} & LPIPS$\downarrow$\;/\;Time\,{\tiny \raisebox{0.3ex}{[\si{\milli\second}]}}  \\
\midrule
Captured \cite{stereolabs2023depthsettings}& 0 & 0.2438\;/\;74.39 & 0.1987\;/\;8.51\\
Monocular \cite{lin2025depthanything3}& 30.47 & 0.2834\;/\;74.39 & 0.2725\;/\;8.51  \\
Multi-view \cite{cao2024mvsformer++}& 40.72 & 0.2412\;/\;74.39 & 0.1976\;/\;8.51 \\
\bottomrule
\end{tabular}
\caption{\label{tbl:ablation_geometry}Quantitative comparison of input depth source (captured stereo~\cite{stereolabs2023depthsettings}, monocular~\cite{lin2025depthanything3} and transformer MVS~\cite{cao2024mvsformer++}) and target depth fusion (ours vs \textit{nvblox}' TSDF~\cite{millane2024nvblox}) on the \textit{LiveNVS} dataset.}
\end{table}%
\begin{figure}[t]\iflatexml\includegraphics[width=\linewidth]{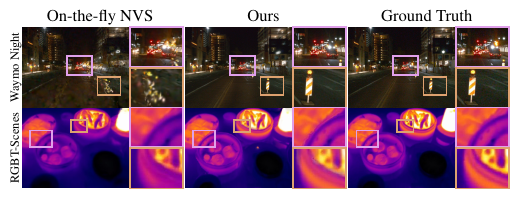}%
\else\par\setbox0\hbox{\kern-5pt\raisebox{-22.07996pt}{\includegraphics{figures_tikz/fig_multimodal.pdf}}\kern-5pt}\ht0=\dimexpr\ht0-5pt\relax\dp0=\dimexpr\dp0-5pt\relax\box0\prevdepth=17.07996pt\relax\fi%
\caption{Thermal images of the \textit{RGBT-Scenes} dataset and automotive scenes captured during night.}%
    \label{fig:multimodal}
\end{figure}%
\section{Discussion and Conclusion}
\fakeparagraph{Dynamic Scenes.} Our point accumulation splits scenes into static and dynamic components.
Likewise, extending the IBR pipeline could enable handling of dynamic scenes.
\fakeparagraph{Accumulation Performance.} The point cloud accumulations' performance depends on size and frequency of depth observations. In our experiments accumulation time was negligible compared to the other parts of the pipeline, hence we consider efficient handling of larger scenes future work.
\fakeparagraph{Edge Artifacts.} Our approach tends to produce artifacts at \textit{occlusion edges}. 
As these stem from misalignment of reconstructed depth maps to ground truth, this invites further research regarding refinement of the depth completion.
\fakeparagraph{Occlusion Resolution.} Our conservative disocclusion resolution allows false negatives. Improving depth maps or the warping process could yield higher image quality.
\fakeparagraph{Data Dependency.} Insufficient sensor coverage results in degradation of image features. The exploration of additional assumptions or heuristics could be considered.
\fakeparagraph{Temporal Stability.} Our weights may cause temporal instability under movement. To reduce these artifacts, we employ temporal anti-aliasing, following~\citet{lewis2025nepo}.
We demonstrate the efficacy in the supp. video (S6).

In summary, we present a live novel-view synthesis approach based on image-based rendering, combining measurement-anchored depth completion with frequency-aware two-band blending. 
We demonstrate its effectiveness across diverse capture settings and heterogeneous sensor modalities.
Visually (LPIPS), our method consistently outperforms all baselines. On PSNR, which is forgiving with low-frequency reconstructions, performance is scene-dependent; still, we improve PSNR by up to 11\,\% relative to the second-best state-of-the-art method for handheld RGB-D captures, and by up to 6\,\% for driving scenarios, while reducing \textit{capture-to-render} time by at least a factor of two.
Despite requiring neither scene-specific optimization nor training, our approach remains competitive on established benchmarks and outperforms prior methods in challenging settings, including night-time and thermal capture, paving the way toward reliable live novel-view synthesis for robust, real-world capture workflows.

\section*{Acknowledgments}
We sincerely thank e.solutions GmbH for supporting this work.
Thanks to Linus Franke for his guidance during early stages of the project.
The authors gratefully acknowledge the scientific support and HPC resources provided by the Erlangen National High Performance Computing Center (NHR@FAU) of the Friedrich-Alexander-Universität Erlangen-Nürnberg (FAU). The hardware is partially funded by the German Research Foundation (DFG).
Noah Lewis was supported by the Bavarian Transformation and Research Foundation AZ-1626-24.

{
    
    \printbibliography                

}

\maketitlesupplementary
\setcounter{section}{0}
\setcounter{equation}{0}
\setcounter{figure}{0}
\setcounter{table}{0}
\setcounter{algocf}{0}
\makeatletter
\renewcommand \thesection{S\@arabic\c@section}
\renewcommand\thetable{S\@arabic\c@table}
\renewcommand \thefigure{S\@arabic\c@figure}
\renewcommand\thealgocf{S\@arabic\c@algocf}
\makeatother

This supplementary material provides supporting methodological information (\secref{sup:sec:method}), further details on baseline (\secref{sup:sec:baselines}) and dataset setup (\secref{sup:sec:datasets}), additional results (\secref{sup:sec:results}) and ablations  (\secref{sup:sec:ablations}) complementing the main paper, and a description of the accompanying video (\secref{sup:sec:video}).
\section{Method}%
\label{sup:sec:method}
\subsection{Depth Completion}%
\label{sup:sec:depthcompletion}

For the high-pass kernel $\kernel$, introduced in the main paper for computing a representative depth value of a pixel's neighborhood, we use the zero-centered $5\times 5$ kernel:
\begin{align}
    \kernel =
     \begin{bsmallmatrix}
        1 & 4 & 6 & 4 & 1\\
        4 & 16 & 24 & 16 & 4\\
        6 & 24 & 0 & 24 & 6\\
        4 & 16 & 24 & 16 & 4\\
        1 & 4 & 6 & 4 & 1
    \end{bsmallmatrix}
\end{align}%
This kernel is applied to $\densedepth^\level$ at each level $\level$ (Algorithm 1:4).
\subsection{SoftEDT}%
\label{sup:sec:softEDT}
\SetKwProg{Fn}{Function}{:}{}
\SetKwFunction{SoftEDT}{SoftEDT}

\newcommand{\Dsq}{D^2}
Our \emph{softEDT} algorithm (Algorithm~\ref{algo:softEDT})
\begin{algorithm*}[tb]\footnotesize
\caption{\label{algo:softEDT}%
  \textsc{SoftEDT}: the function calculates a squared version of our approximate, smoothed Euclidean distance
  transform; a final square-root operation yields (approximate) linear distances.
  Here, $\boxplus$ denotes min-plus convolution with the $3\times3$ kernel
  $\Delta={\tiny\left[\begin{smallmatrix}2&1&2\\1&0&1\\2&1&2\end{smallmatrix}\right]}$ of
  squared neighbour distances, $\operatorname{pool}_{\min}$ is $2\times2$
  minimum pooling, and $\operatorname{up}(\,\cdot\,;w,h)$ is bilinear upsampling to
  size $w\times h$. (We indicate the storage of \emph{squared} distances by $D^2$, not to be mistaken for an actual square operation.)}
\KwIn{field $\Dsq\in(\mathbb{R}_{\ge0}\cup\{\infty\})^{w\times h}$, foreground $0$, background $\infty$}
\KwOut{approximate squared distance to the foreground}
\Fn{\SoftEDT{$\Dsq$}}{
  $\Dsq \gets \Dsq \boxplus \Delta$ \tcp*{min-plus convolution with the $3\times3$ parabolic kernel $\Delta$}
  \If{$\max(w,h) > 2$}{
    $\Dsq_{\downarrow} \gets \tfrac18 + \tfrac14\,\operatorname{pool}_{\min}(\Dsq)$ \tcp*{min-pool, octave rescale, conservative half-pixel offset}
    $\Dsq_{\downarrow} \gets$ \SoftEDT{$\Dsq_{\downarrow}$}\;
    $\Dsq \gets \min\!\big(\Dsq,\; 4\cdot\operatorname{up}(\Dsq_{\downarrow};\,w,h)\big)$ \tcp*{bilinear upsample, octave rescale}
  }
  \Return $D^2$\;
}
\end{algorithm*}
computes an approximate, smoothed Euclidean distance transform (EDT) $D$
that stays faithful near the foreground and becomes increasingly smoothed away from
it. A squared distance field $\Dsq$ (foreground $\rightarrow 0$, background $\rightarrow +\infty$), is refined in a multi-resolution approach. At each scale, squared distances propagate by a single min-plus convolution
$(\Dsq\boxplus \Delta)(\mathbf{p})=\min_{\mathbf{q}} \Dsq(\mathbf{q})+\Delta(\mathbf{p}-\mathbf{q})$
over a $3\times3$ paraboloid, $\Delta=\tiny\left[\begin{smallmatrix}2&1&2\\1&0&1\\2&1&2\end{smallmatrix}\right]$
of squared distances to its eight neighbors. Since one pass propagates the distance field by a pixel, longer ranges stem from recursion: the field is $2\times2$ min-pooled, biased by a conservative offset of
$\bigl(\frac{\sqrt{1/2}}{2}\bigr)^2 = \frac{1}{8}$,
and rescaled into the respective coarser distance units, and then further transformed recursively,
before bilinear upsampling that coarser with current levels by pointwise minimum. Linearized approximate distances are finally determined as $D=\sqrt{\Dsq}$.
\section{Baselines}
\label{sup:sec:baselines}
In the following, we give further detail on the employed baselines and our experimental setup of their methods.

For \textit{On-the-fly NVS}~\cite{meuleman2025onthefly}, we report performance by measuring the average processing time of a single frame in a sequence, including Gaussian sampling and optimization.
As this quantifies the time interval between the last train view detected and the rendering of the test view, we argue that this is the most analogous to \textit{capture-to-image} time.
Although we pass ground truth poses to the method, they still perform their pose initialization, which needs eight keyframes by default. 
They select keyframes based on similarity to previous keyframes to choose sufficiently differing frames.
In some scenes, this led to their approach not being capable of operating under the experiments' constraints, as a camera stream from a front facing automotive camera does not change significantly between frames.
We modified their pipeline such that every frame is accepted as a keyframe to enable fair comparison.
Notably, their results improved with our modifications in place, compared to their standard keyframe selection for our datasets.

Image-based rendering approaches, like \textit{Inovis} by \citet{harrerfranke2023inovis} and \textit{LiveNVS} by \citet{fink2023livenvs}, are compared as follows. %
We compare to \textit{Inovis} on all live reconstruction datasets. For all comparisons, we use the checkpoint published with their codebase that has been trained on their custom office dataset and finetuned on Kitti 360 data~\cite{liao2022kitti}. %
As their method assumes point data to be colored, we color input data using the input images.
\textit{LiveNVS} is evaluated exclusively on their own dataset, as other datasets do not provide dense depth images, which the method depends on.
Another relevant approach is \textit{Quark} presented by \citet{flynn2024quark}, who report high rendering quality and efficiency. However, a direct comparison is not feasible due to the closed-source nature of their approach.

As representatives for feed-forward Gaussian reconstruction approaches, we compare to \textit{DepthAnythingV3} by \citet{lin2025depthanything3} and \textit{Long-LRM} by \citet{ziwen2025longlrm}.
Due to VRAM demands, we keep \textit{DepthAnythingV3}'s default processing resolution, which sets the width to 504 and scales the height according to the input aspect ratio (processing in native dataset resolution imposed VRAM requirements $>$24~GB).
When passing ground truth poses into the model, predicted and ground truth poses are aligned by a least squares estimation~\cite{umeyama1991least}.
We found this to cause severe misalignments and simplified the alignment strategy to only apply the scale transformation between the closest input view (i.e the pose captured right before the test view) and the test view. This significantly improved metrics for the NVS use case.
For \textit{Long-LRM}, we utilized the author's \textit{Self-Reimplemented Version}. The provided checkpoint requires images of height 544 and aspect ratio $\leq$ 1.
Hence, input images are rescaled to match the desired height. 
For input images with aspect ratio $<$ 1 (\textit{Argoverse 2} dataset), we utilize their option to rescale images to the max input resolution of 960$\times$544.
While images are downscaled for reconstruction, the feed-forward predicted 3D gaussians are rendered in the target frame's resolution.
\begin{table*}[t]
\centering
\begin{tabular}{l|c|c|c}
\toprule
 Dataset & Proc. Res. & Sensor Setup & Scene Type \\
\midrule
\textit{Waymo~\cite{sun2020waymo}} & 1280p & LiDAR\,+\,5 cameras & automotive \\
\textit{nuScenes~\cite{caesar2020nuscenes}} & 900p & LiDAR\,+\,6 cameras & automotive \\
\textit{Argoverse 2~\cite{wilson2021argoverse2}} & 1550p & LiDAR\,+\,7 cameras & automotive \\
\textit{LiveNVS~\cite{fink2023livenvs}} & 720p\,/\,1080p & RGB-D & handheld\,(+\,SLAM) \\
\textit{MipNerf 360~\cite{barron2022mipnerf360}} & 520p\,/\,820p & RGB & handheld\,(+\,COLMAP) \\
\textit{Tanks and Temples~\cite{Knapitsch2017}} & 1080p & RGB & handheld\,(+\,COLMAP) \\
\textit{RGBT-Scenes~\cite{lu2024thermalgaussian}} & 480p & RGB\,+\,Thermal & thermal\,(+\,COLMAP) \\
\bottomrule
\end{tabular}
\caption{\label{sup:tbl:dataset_descr}Specification of the benchmark datasets, detailing processing resolution and sensor setup across various scene types.}
\end{table*}
\begin{table*}[!ht]
\centering
\begin{tabular}{l|l}
\toprule
 \multicolumn{1}{c}{\textit{Waymo Open Dataset~\cite{sun2020waymo}}} & \multicolumn{1}{c}{\textit{Argoverse 2~\cite{wilson2021argoverse2}}} \\
\midrule

\texttt{2601205676330128831\_4880\_000\_4900\_000} & \texttt{00a6ffc1-6ce9-3bc3-a060-6006e9893a1a} \\
\texttt{4140965781175793864\_460\_000\_480\_000} & \texttt{0a132537-3aec-35bb-af13-7faa0811000d} \\
\texttt{7247823803417339098\_2320\_000\_2340\_000} & \texttt{0b324587-6097-3f92-a07a-a44f48c85d9e} \\
\texttt{8688567562597583972\_940\_000\_960\_000} & \texttt{01bb304d-7bd8-35f8-bbef-7086b688e35e} \\
\texttt{8993680275027614595\_2520\_000\_2540\_000} & \texttt{03b2cf2d-fb61-36fe-936f-36bbf197a8ac} \\
\texttt{10504764403039842352\_460\_000\_480\_000} & \texttt{03fba633-8085-30bc-b675-687a715536ac} \\
\texttt{11987368976578218644\_1340\_000\_1360\_000} & \texttt{022af476-9937-3e70-be52-f65420d52703} \\
\texttt{14188689528137485670\_2660\_000\_2680\_000} & \texttt{04973bcf-fc64-367c-9642-6d6c5f363b61} \\
\texttt{14737335824319407706\_1980\_000\_2000\_000} & \\
\texttt{15272375112495403395\_620\_000\_640\_000} & \\
\texttt{17792628511034220885\_2360\_000\_2380\_000} & \\
\bottomrule
\end{tabular}
\caption{\label{sup:tbl:waymoargoscenes}List of scenes chosen from \textit{Waymo Open Dataset} and \textit{Argoverse 2}. Presented scenes show comparatively reduced dynamic content.}
\end{table*}

The \textit{DrivingForward}~\cite{tian2025drivingforward} checkpoint available with their codebase was trained on the nuScenes~\cite{caesar2020nuscenes} training data split exclusively,
implying a potential domain gap when evaluated on other automotive datasets. 
Although this domain gap may contribute to reduced performance, the approach still generalized as expected and produced meaningful results on other datasets.
We compute their results on the default resolution of $640\times 352$, since increasing their resolution led to worse metrics in our experiments.
Their implementation supports either three or six input cameras.%
\section{Datasets}%
\label{sup:sec:datasets}
All datasets used in our experiments are listed in \tblref{sup:tbl:dataset_descr}. In the context of live reconstruction, we employ timestamped datasets, thus allowing to simulate an updating scene state.
For each test frame, a slice containing the data of $n$ preceding seconds is used as input for any method.
Therefore, the test frame represents an extrapolation to the next novel viewpoint, rather than an interpolation between images that have already been captured.

\fakeparagraph{Automotive.}
As mentioned in the paper, we evaluate on automotive datasets \textit{Waymo Open Dataset}~\cite{sun2020waymo}, \textit{nuScenes}~\cite{caesar2020nuscenes} (10 scenes), and \textit{Argoverse 2}~\cite{wilson2021argoverse2} (8 scences).
For \textit{nuScenes}, their \textit{mini} subset containing 10 scenes is utilized.
From \textit{Waymo Open Dataset} and \textit{Argoverse 2} we choose eleven and eight scenes, respectively.
The utilized scenes exhibit comparatively reduced dynamic content, given that all evaluated methods focus on static scenes.
See \tblref{sup:tbl:waymoargoscenes} for a detailed list of scenes used from \textit{Waymo} and \textit{Argoverse 2}.
To enable processing with methods that are not capable of managing multi-camera scenarios (i.e. \textit{On-the-fly NVS} and \textit{Inovis}), we evaluate all methods on the front facing cameras only.
Since \textit{DrivingForward} is trained for the \textit{nuScenes} dataset comprising a multi-camera setup, we provide all six cameras for nuScenes for this approach.
From \textit{Waymo Open Dataset}'s five and \textit{Argoverse 2}'s seven cameras, we provide the three front-facing cameras to \textit{DrivingForward} to enable a fair comparison.
Although our method is capable of handling multiple cameras, we limit our method in the evaluations as well.
The only exceptions for this are the \textit{Waymo Open Dataset} scene in Figure~8 and the accompanying video (see \secref{sup:sec:video}), where we utilize all cameras of datasets in question.
\newpage
For all automotive scenes, every 32nd frame is designated as a test frame. 
The data from the previous $\numsomething=5$ seconds is considered to be input for each test frame. 
Experiments are conducted on the native dataset resolution, except for \textit{Argoverse 2}, where we crop the front camera image to remove the bonnet ($2048 \times 1550 \rightarrow 1800\times 1550$).
The absence of depth images in automotive datasets renders a comparison to  \textit{LiveNVS} infeasible. 
\fakeparagraph{RGB-D.} The keyframed version of the \textit{LiveNVS} dataset~\cite{fink2023livenvs} comprises $\sim\!1$ image per second.
We thus choose a larger timing window of $\numsomething=15$, to ensure the acquisition of sufficient images for each approach. 
Each 16th frame is designated as a test frame.
The dataset was requested from the authors.
\fakeparagraph{Offline.} In the context of offline scenarios, we deploy a common NVS evaluation by reporting metrics on a test set comprising every eighth image.
Seven scenes of the \textit{MipNerf 360}~\cite{barron2022mipnerf360} dataset (i.e. garden, bicycle, bonsai, counter, kitchen, room, and stump) and four scenes of the \textit{Tanks and Temples}~\cite{Knapitsch2017} dataset (i.e. caterpillar, francis, playground, and train) are utilized. A dense COLMAP~\cite{schonberger2016structure} reconstruction is performed before evaluation. We deploy \textit{MipNerf 360} on quarter resolution and \textit{Tanks and Temples} on full resolution.
\fakeparagraph{Thermal.}
The versatility of our method across different modalities is validated by a comparison of the quality of synthesis on thermal images.
We compare our approach on the \textit{RGBT-Scenes} dataset presented by~\citet{lu2024thermalgaussian}.
The evaluation of this dataset is analogous to that of the previously described offline scenes.
\section{Additional Results}%
\label{sup:sec:results}
\begin{figure*}[t]\iflatexml\includegraphics[width=\linewidth]{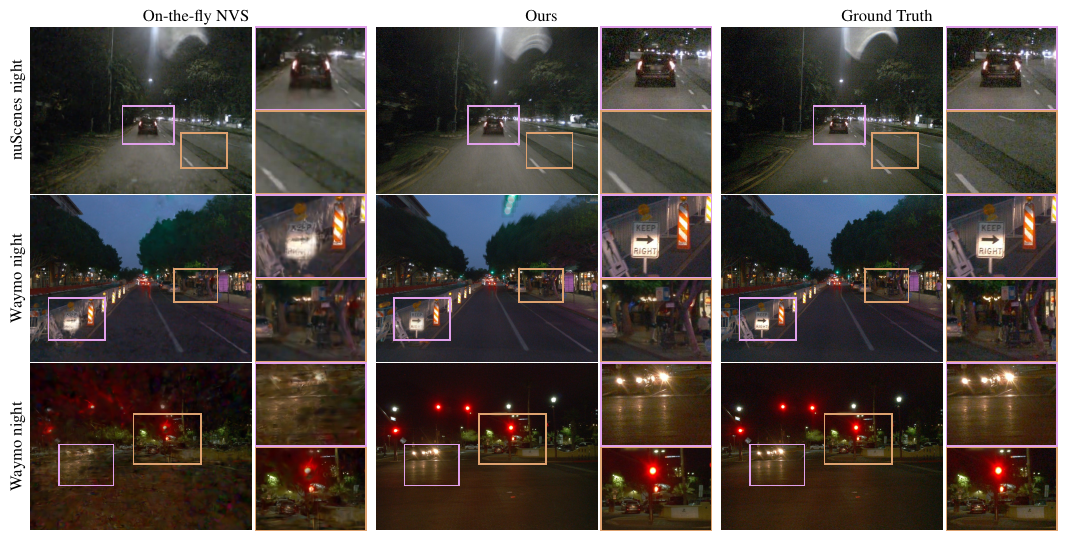}%
\else\par\setbox0\hbox{\kern-5pt\raisebox{-42.81458pt}{\includegraphics{figures_tikz/sup_fig_results_multimodal_night.pdf}}\kern-5pt}\ht0=\dimexpr\ht0-5pt\relax\dp0=\dimexpr\dp0-5pt\relax\box0\prevdepth=37.81458pt\relax\fi%
\caption{Live NVS results on automotive scenes captured during night.}%
    \label{sup:fig:results_multimodal_night}
\end{figure*}%

\begin{figure*}[p]\iflatexml\includegraphics[width=\linewidth]{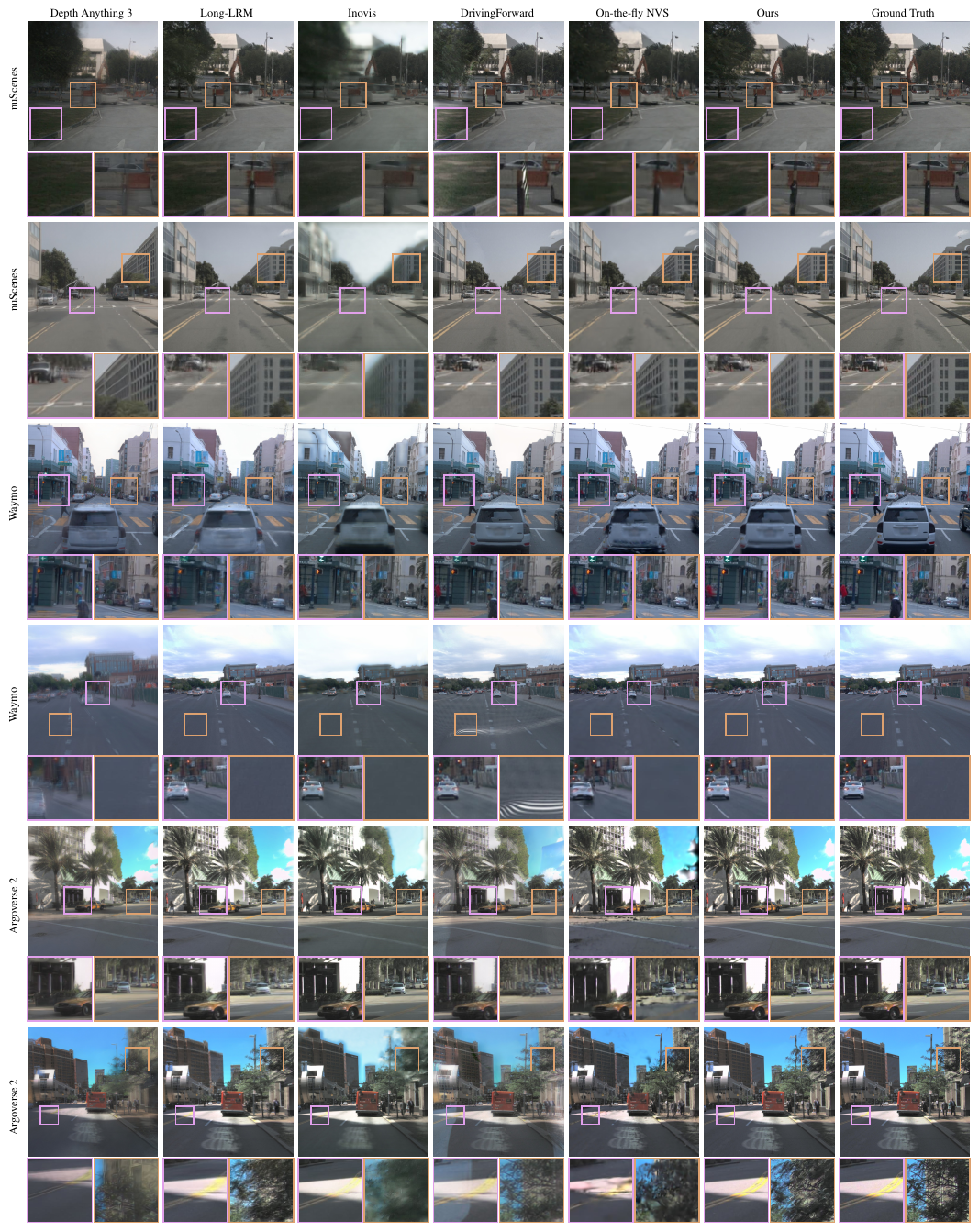}%
\else\par\setbox0\hbox{\kern-5pt\raisebox{-53.92613pt}{\includegraphics{figures_tikz/sup_fig_results_live_automotive.pdf}}\kern-5pt}\ht0=\dimexpr\ht0-5pt\relax\dp0=\dimexpr\dp0-5pt\relax\box0\prevdepth=48.92613pt\relax\fi%
\caption{Qualitative comparison in automotive live scenarios form the \textit{nuScenes}, \textit{Waymo Open Dataset}, and \textit{Argoverse 2} datasets.}%
    \label{sup:fig:results_live_automotive}
\end{figure*}%

\begin{figure*}[p]\iflatexml\includegraphics[width=\linewidth]{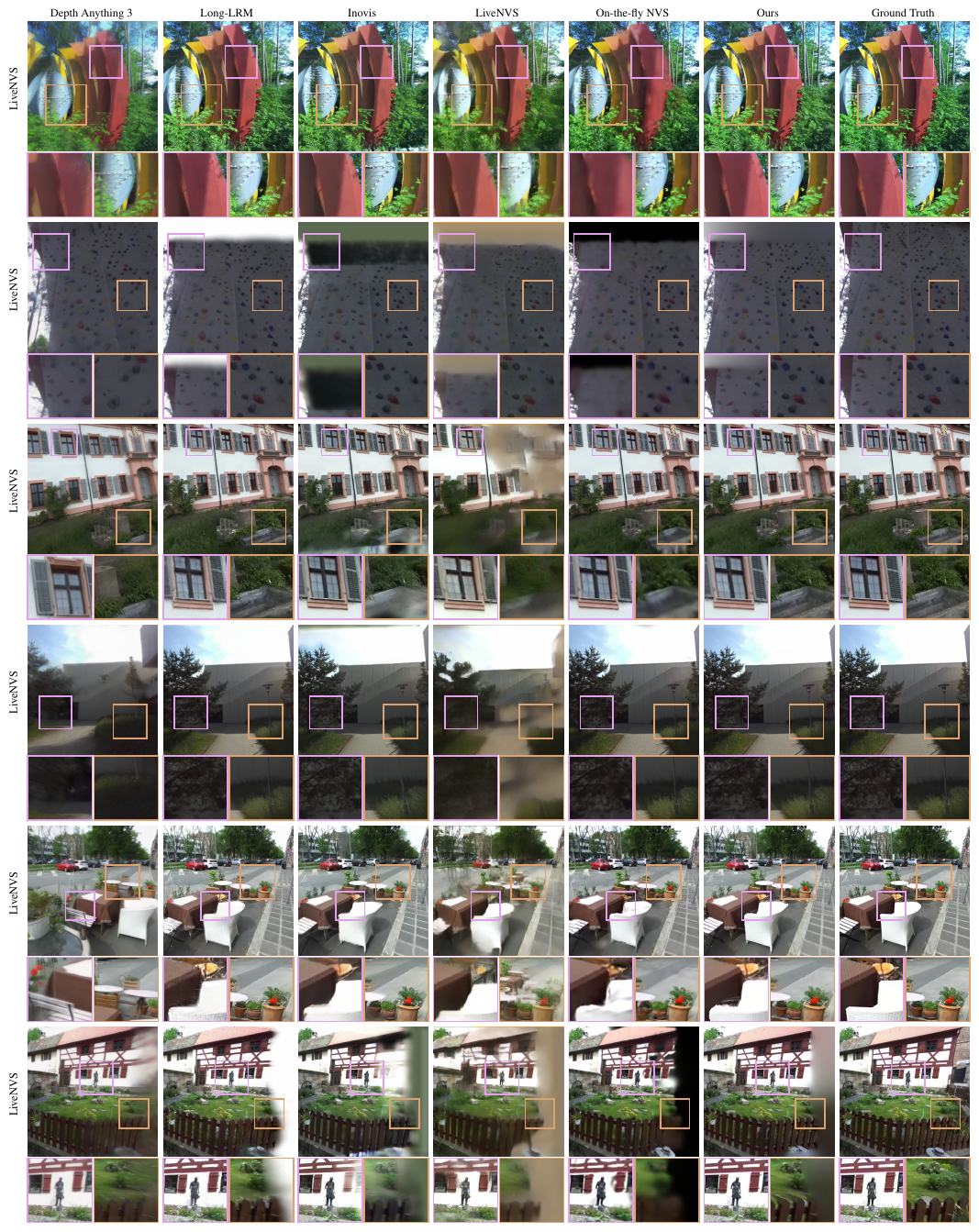}%
\else\par\setbox0\hbox{\kern-5pt\raisebox{-53.9336pt}{\includegraphics{figures_tikz/sup_fig_results_live_handheld.pdf}}\kern-5pt}\ht0=\dimexpr\ht0-5pt\relax\dp0=\dimexpr\dp0-5pt\relax\box0\prevdepth=48.9336pt\relax\fi%
\caption{Qualitative comparison in handheld live scenarios form the \textit{LiveNVS} dataset.}%
    \label{sup:fig:results_live_handheld}
\end{figure*}%

\begin{figure*}[t]\iflatexml\includegraphics[width=\linewidth]{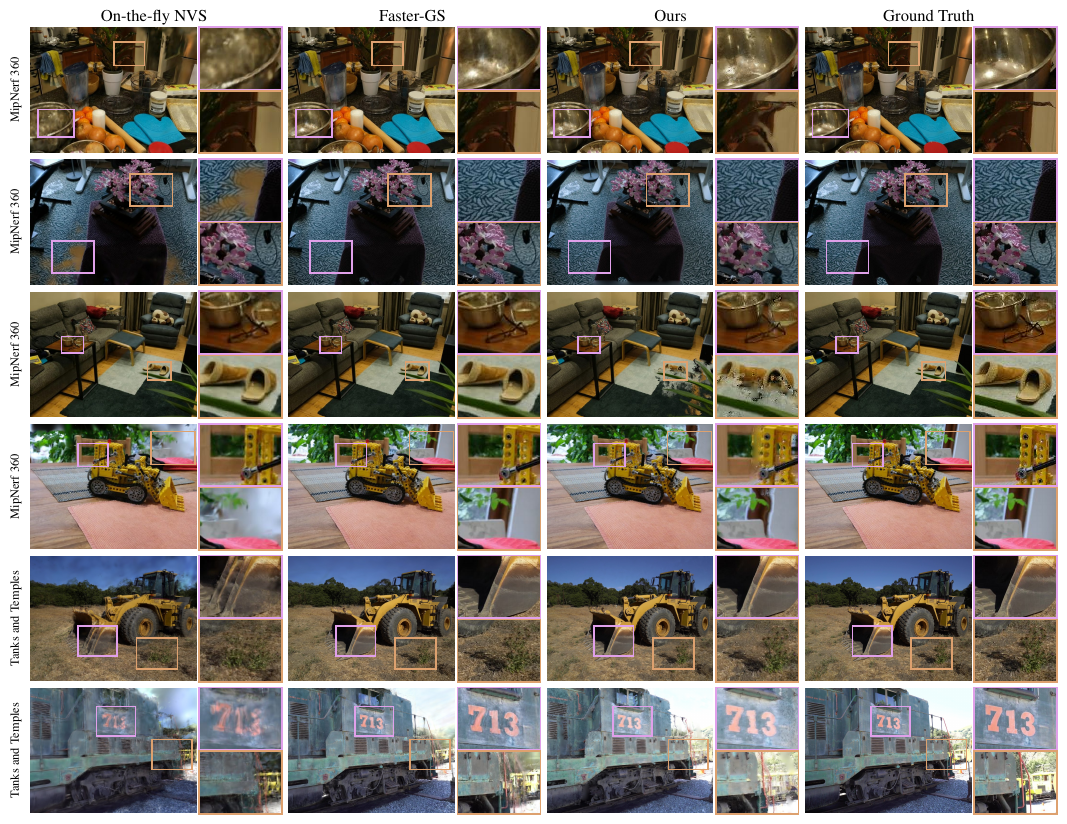}%
\else\par\setbox0\hbox{\kern-5pt\raisebox{-32.92842pt}{\includegraphics{figures_tikz/sup_fig_results_offline.pdf}}\kern-5pt}\ht0=\dimexpr\ht0-5pt\relax\dp0=\dimexpr\dp0-5pt\relax\box0\prevdepth=27.92842pt\relax\fi%
\caption{Qualitative comparison in offline scenarios. Four \textit{MipNerf 360} and two \textit{Tanks and Temples} scenes are displayed.}%
    \label{sup:fig:results_offline}
\end{figure*}%

\begin{figure*}[t]\iflatexml\includegraphics[width=\linewidth]{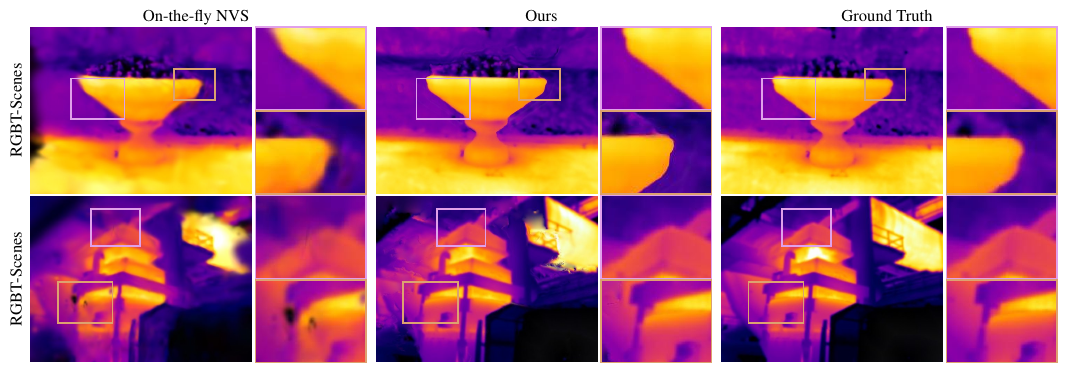}%
\else\par\setbox0\hbox{\kern-5pt\raisebox{-87.96832pt}{\includegraphics{figures_tikz/sup_fig_results_multimodal.pdf}}\kern-5pt}\ht0=\dimexpr\ht0-5pt\relax\dp0=\dimexpr\dp0-5pt\relax\box0\prevdepth=82.96832pt\relax\fi%
\caption{Thermal images of the \textit{RGBT-Scenes} dataset.}%
    \label{sup:fig:results_multimodal}
\end{figure*}%

Supplementing the results displayed in the manuscript, we provide further qualitative results.
\figref{sup:fig:results_live_automotive} presents more comparisons in live automotive settings while \figref{sup:fig:results_live_handheld} display further comparisons on the handheld RGB-D \textit{LiveNVS} dataset.
Here, we provide visual comparisons with methods evaluated only quantitatively in the main manuscript.
\figref{sup:fig:results_offline} extends comparisons in offline scenarios and \figref{sup:fig:results_multimodal_night} and \figref{sup:fig:results_multimodal} provide further examples of low-light and thermal scenarios.
\begin{figure*}[t]\iflatexml\includegraphics[width=\linewidth]{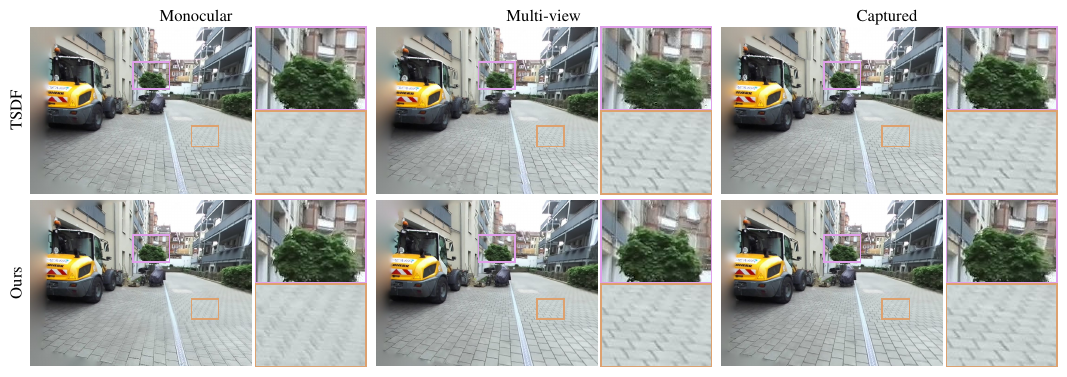}%
\else\par\setbox0\hbox{\kern-5pt\raisebox{-42.73383pt}{\includegraphics{figures_tikz/sup_fig_ablation_geometry.pdf}}\kern-5pt}\ht0=\dimexpr\ht0-5pt\relax\dp0=\dimexpr\dp0-5pt\relax\box0\prevdepth=37.73383pt\relax\fi%
\caption{Ablation of source of geometry. Comparison of input depth acquisition techniques (captured RGB-D from Zed 2i camera~\cite{stereolabs2023depthsettings}, monocular depth estimation~\cite{lin2025depthanything3}, transformer-based multi-view depth estimation~\cite{cao2024mvsformer++}) and target depth fusion (our depth completion vs \textit{nvblox} TSDF~\cite{millane2024nvblox}) }%
    \label{sup:fig:ablation_geometry}
\end{figure*}%

\begin{figure}[t]
    \centering
    \begin{minipage}{0.495\linewidth}
    \centering {\small Ours + \textit{MVS-Former++} \vphantom{Gg}}
    \end{minipage}\hfill%
    \begin{minipage}{0.495\linewidth}
    \centering {\small  Ours using LiDAR data\vphantom{Gg}}
    \end{minipage}\\%
    \includegraphics[width=0.48\linewidth]{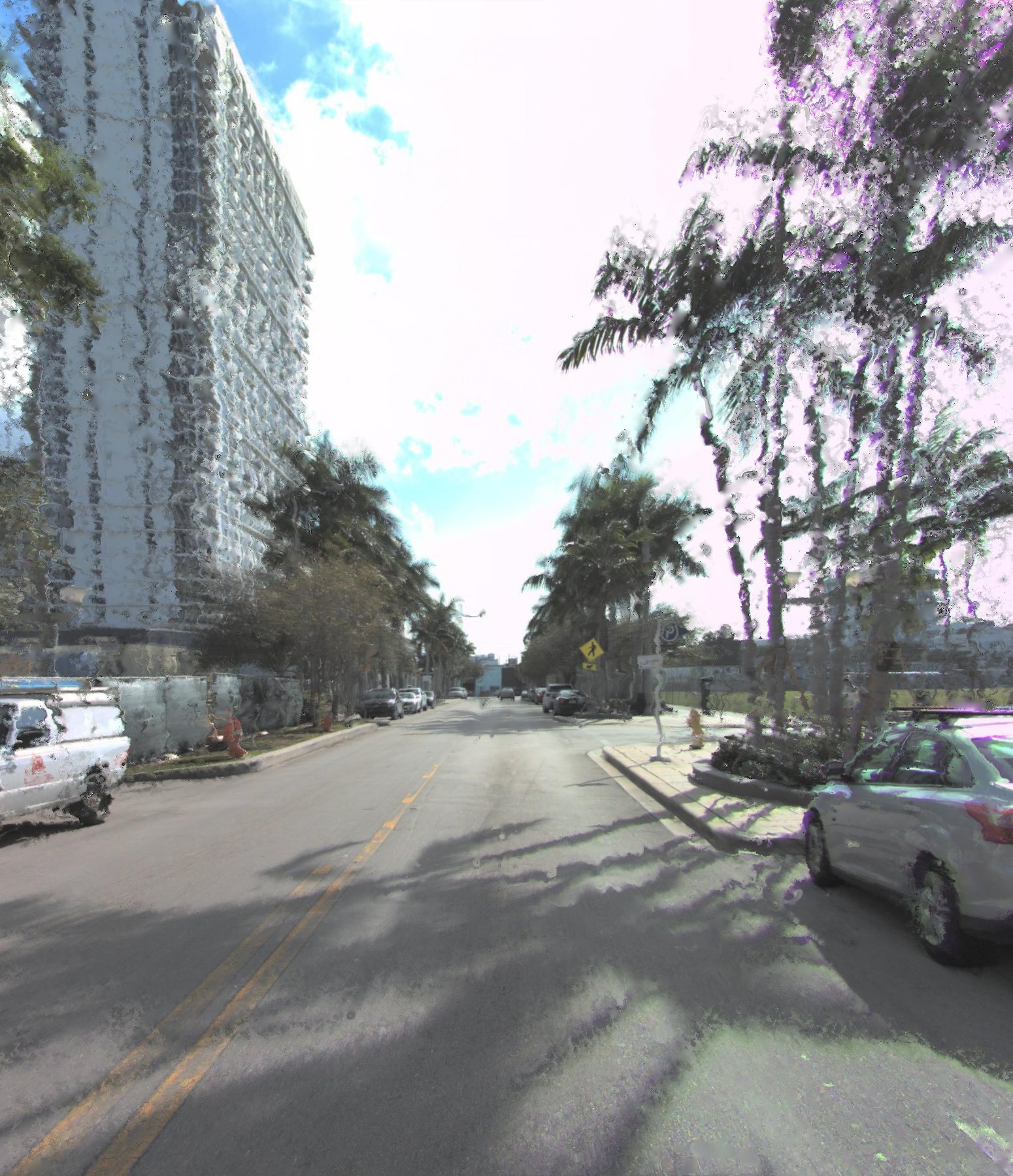}%
    \hfill
    \includegraphics[width=0.48\linewidth]{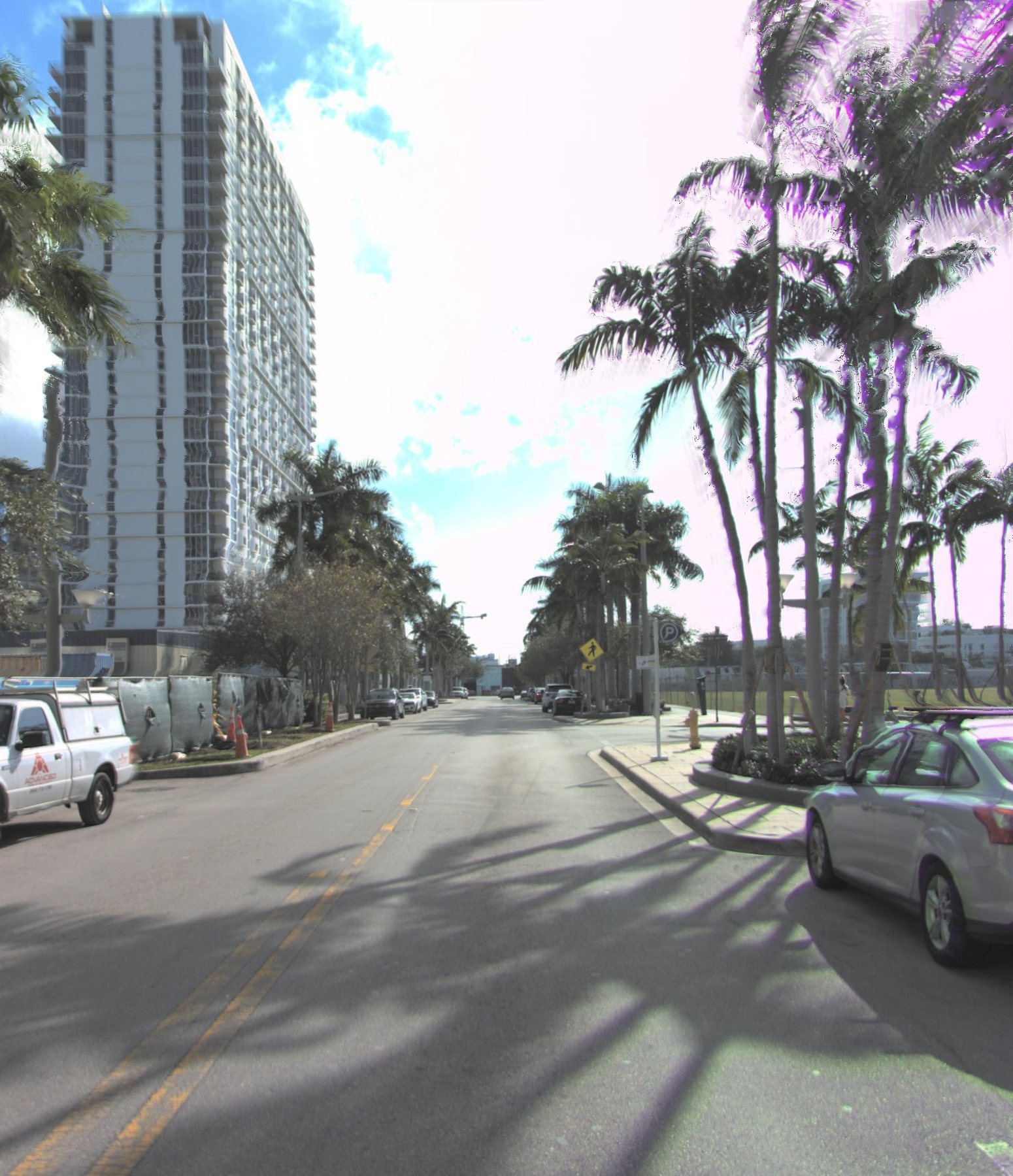}%
     
    \caption{Failure case of multi-view depth estimation with \textit{MVSFormer++}. In this automotive example including a strictly forward trajectory without lateral baseline, the depth estimation produces lots of floaters, severely impacting image quality (\textit{left}), while using LiDAR data as depth data reveals no such issues (\textit{right}).}%
    \label{sup:fig:ablation_geometry_2}
\end{figure}%

\subsection{Discussion of Metrics: PSNR vs. LPIPS}
In comparison to related work, our approach performs notably better in terms of the LPIPS metric than the PSNR metric. 
LPIPS aligns better with human perception as shown in the original LPIPS paper by \citet{zhang2018unreasonable} and \citet{zuo2022view}.
Compared to the PSNR metric, LPIPS is more robust towards distortions, noise, blur, and shifts.
\citet{zuo2022view} specifically highlighted the severely negative impact of slight misalignment on PSNR scores.
As discussed, the artifacts most prominent with our approach are due to misalignment of reconstructed depth with ground truth, which shows itself as a visible seam in the area of depth discontinuities.
These seams are visually similar to image shifts, hurting PSNR scores severely.
Furthermore, LPIPS rightfully penalizes blurry images with a worse score, while PSNR tends to increase when blurring an image.
As gaussians are blurry primitives by definition, 3D gaussian splatting based approaches perform well with PSNR scores.
We argue that this results in comparatively worse PSNR scores, while our approach consistently outperforms competitors in LPIPS, which aligns better with human perception.

\section{Further Ablations}%
\label{sup:sec:ablations}
\fakeparagraph{Ablation:\;Depth Source (continued).}
To ablate the source of geometric information, two different tasks are distinguished: acquisition of depth data for each input view and the fusion of input view depth to derive a depth in target view. These tasks differ significantly, as for the former, RGB data is available, while for the latter it is not. 

For per input view depth, our method leverages measured depth observations from LiDAR or RGB-D sensors. In the absence of such data, per input view depth can be derived from posed RGB images. Therefore, we compare captured RBG-D stero depth of the \textit{LiveNVS}~\cite{fink2023livenvs} dataset (Zed 2i camera~\cite{stereolabs2023depthsettings}) to two neural depth estimation techniques. As a metric monocular depth estimator, we employ \textit{Depth Anything 3}'s \texttt{DA3METRIC-LARGE} checkpoint.
For multi-view depth estimation, we pass input images and poses to \textit{MVSFormer++}~\cite{cao2024mvsformer++}.
For the acquisition of target view depth, we compare our point-based depth completion to the open source TSDF of the \textit{nvblox} framework, which is a faster open source variant compared to the original TSDF~\cite{curless1996volumetric}. We use a voxel size of 7.5 cm, which acts as a tradeoff between computation times and depth quality,

Quantitative results can be found in \tblref{tbl:ablation_geometry} while supplemental~\figref{sup:fig:ablation_geometry} extends the evaluation with a qualitative comparison. The results demonstrate the robustness and performance of our approach. Monocular metric depth estimation leads to significant ghosting artifacts due to depth values that are not multi-view consistent. \textit{MVSFormer++} depth leads to similar metrics as captured depth, while the computational cost of 40.72 ms per input view can have a significant impact on the systems performance. This demonstrates our systems flexibility to be deployed in scenarios without depth sensors, albeit with a slight performance penalty. Additionally, \textit{MVSFormer++} predictons may deteriorate when the capture path follows, e.g., a constant forward trajectory with no lateral baseline, see the automotive example in \figref{sup:fig:ablation_geometry_2}.
In terms of target view depth fusion, our approach outperforms TSDF in both, image quality and processing speed.

\begin{figure*}[t]\iflatexml\includegraphics[width=\linewidth]{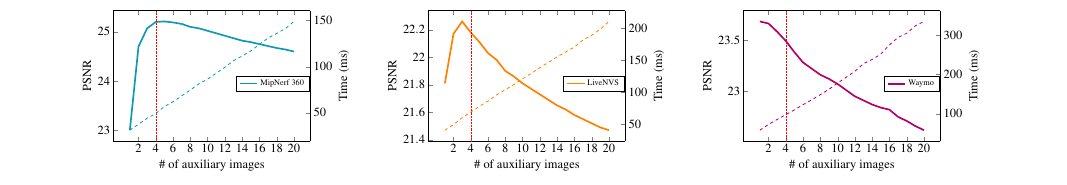}%
\else\par\setbox0\hbox{\kern-5pt\raisebox{-5.0pt}{\includegraphics{figures_tikz/sup_fig_ablation_numimages.pdf}}\kern-5pt}\ht0=\dimexpr\ht0-5pt\relax\dp0=\dimexpr\dp0-5pt\relax\box0\prevdepth=0.0pt\relax\fi%
\caption{\label{sup:fig:ablation_numimages} PSNR and render time against number of used auxiliary images $\numauxiliary$. The red dotted line illustrates the chosen trade-off $\numauxiliary =4$.}
\end{figure*}

\begin{figure}[t]\iflatexml\includegraphics[width=\linewidth]{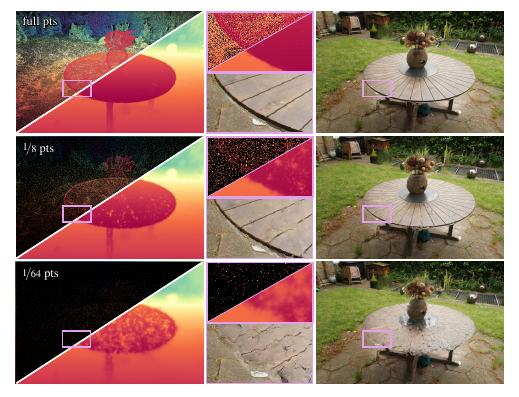}%
\else\par\setbox0\hbox{\kern-5pt\raisebox{-93.27666pt}{\includegraphics{figures_tikz/fig_numberofpoints.pdf}}\kern-5pt}\ht0=\dimexpr\ht0-5pt\relax\dp0=\dimexpr\dp0-5pt\relax\box0\prevdepth=88.27666pt\relax\fi%
\caption{\label{fig:numberofpoints}%
           Render quality with varying number of points on MipNerf 360~\cite{barron2022mipnerf360} garden scene.  Sparse depth maps, the respective dense depth maps, the rendering result, and ground truth are displayed under utilization of varying subsampling factors of points. Full point cloud contains 3.1\,mio points.
           }
\end{figure}

\fakeparagraph{Ablation:\;Number of Points.}
We show the performance of the depth completion by comparing its performance under a varying number of points, illustrated in \figref{fig:numberofpoints}.
While rendering quality is highest with all points utilized, it drops insignificantly when only using one eighth of the points. Quality starts degrading more significantly when using $\nicefrac{1}{64}$ points compared to the original COLMAP~\cite{schonberger2016structure} reconstruction. 
This demonstrates the robustness of our approach towards the absence of a specifically dense point cloud.
\fakeparagraph{Ablation:\;Number of Auxiliary Images.}
\label{sec:ablation_numimages.}
We ablate the number of auxiliary images~ $\numauxiliary$.
While render time scales linearly with the number of auxiliary images $\numauxiliary$, PSNR peaks at one to four images. 
Since using a single image yields lacking coverage of image data in the novel view, especially when deviating further from the capture trajectory, 
we choose $\numauxiliary=4$ as a tradeoff between coverage, quality, and performance (see \figref{sup:fig:ablation_numimages}).
\begin{figure}[t]\iflatexml\includegraphics[width=\linewidth]{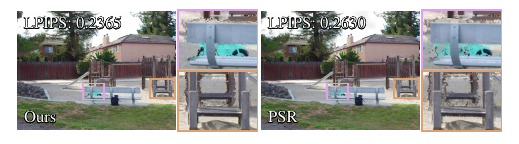}%
\else\par\setbox0\hbox{\kern-5pt\raisebox{-32.20409pt}{\includegraphics{figures_tikz/fig_ablation_psr.pdf}}\kern-5pt}\ht0=\dimexpr\ht0-5pt\relax\dp0=\dimexpr\dp0-5pt\relax\box0\prevdepth=27.20409pt\relax\fi%
\caption{Comparison of our pipeline with the depth completion replaced with the Point sample rendering~\cite{grossman1998point} approach.}
    \label{fig:ablation_psr}
\end{figure}
\fakeparagraph{Ablation:\;Depth Completion.} We compare the performance of our depth completion with the one by~\citet{grossman1998point}, by incorporating their approach into our pipeline. %
\figref{fig:ablation_psr} demonstrates how the proposed method improves both, metrics and visuals, specifically at depth discontinuities. %
\fakeparagraph{Ablation:\;Blending.}
To assess the quality of our \emph{softEDT}-based blending weights, we emplace the learned blending weight prediction of \textit{Deep Blending}~\cite{hedman2018deep} into our IBR pipeline.
As input, we use our warped candidate images~$\setof{\inspace{\generalimage}{\spacetargetview}}$. In contrast to our method, their network expects densely populated inputs, i.e., no holes in input images.
Conversely, we employ the previously mentioned \textit{PushPull}~\cite{gortler1996lumigraph} algorithm to fill any holes arising from unseen geometry in the warping stage.
The network directly predicts a novel view. See \tblref{tbl:ablation_deepblending} for a quantitative and \figref{fig:ablation_deepblending} for a qualitative evaluation on the \textit{LiveNVS} dataset. 
Our blending is tailored towards emphasizing image data far away from \textit{occlusion edges} and subsequently shows better results in these areas. \textit{Deep Blending}'s network tends produce oversmoothed results, while also entailing a higher runtime cost.

\fakeparagraph{Ablation:\;Two-band Blending.}
The impact of deploying a multi frequency approach for blending can be observed in~Fig. 3.%
\setlength{\intextsep}{8pt plus 2pt minus 2pt}%
\setlength{\columnsep}{6pt}%
\begin{wrapfigure}{r}{0.5\linewidth}%
    \centering%
    \begin{tikzpicture}[scale=0.4]
    \tikzstyle{every node}=[font=\Large]
    \begin{axis}[
        xlabel={$\sigma$\vphantom{Gg}},
        ylabel={LPIPS},
        ylabel near ticks,
        width=8.25cm,
        height=6cm,
        xtick={0, 0.5, 0.95, 1.5, 2.5},
        ymajorgrids=true,
        grid=none,
        scaled ticks=false,
        style={/pgf/number format/fixed, /pgf/number format/precision=3},
        ]
        \addplot [
        color=twgreen, style={ultra thick}
        ]
        table {data/data_ablation_blending_sigma.txt};
        \draw[red,densely dotted] (axis cs:0.95,\pgfkeysvalueof{/pgfplots/ymin}) -- (axis cs:0.95,\pgfkeysvalueof{/pgfplots/ymax});
    \end{axis}
    \end{tikzpicture}\\[-1.681ex]
\end{wrapfigure}%
\setlength{\columnsep}{10pt}%
Additionally, we ablate the impact of $\sigma$ utilized in the frequency separation's gauss kernel for our adaption of two-band blending~\cite{brown2003recognising}. %
$\sigma = 0$ is equivalent to a simple weighted sum, as the separation results in the input image being assigned to the low band exclusively.
Two-band blending improves LPIPS compared to a simple weighted sum. 

\begin{figure*}[t]\iflatexml\includegraphics[width=\linewidth]{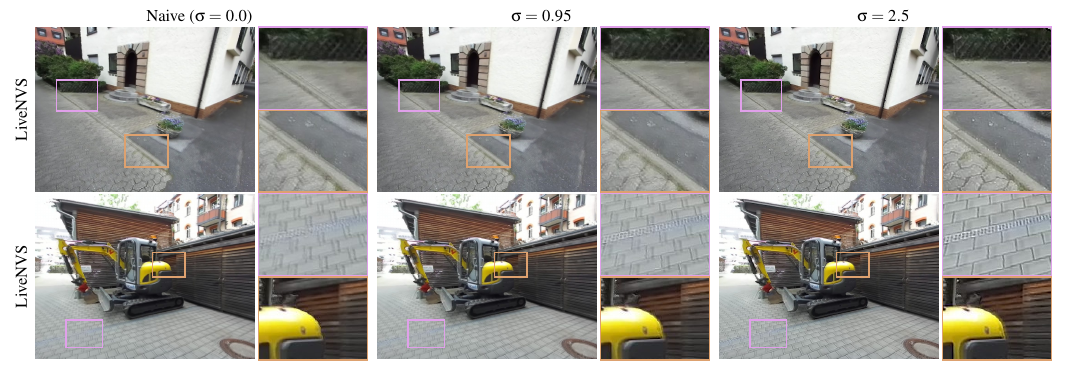}%
\else\par\setbox0\hbox{\kern-5pt\raisebox{-88.45584pt}{\includegraphics{figures_tikz/fig_jittered.pdf}}\kern-5pt}\ht0=\dimexpr\ht0-5pt\relax\dp0=\dimexpr\dp0-5pt\relax\box0\prevdepth=83.45584pt\relax\fi%
\caption{Varying $\sigma$ in presence of calibration errors. Input poses have been artificially jittered, to the point where the naive approach and also our proposed $\sigma = 0.95$ is not able to prevent ghosting artifacts. Increasing $\sigma$ in this scenario leads to high quality images despite calibration errors being present. }%
    \label{fig:jittered}
\end{figure*}

\begin{figure}[!t]\iflatexml\includegraphics[width=\linewidth]{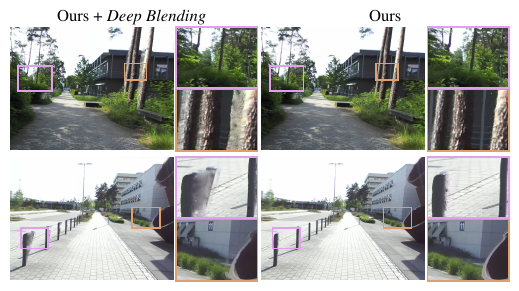}%
\else\par\setbox0\hbox{\kern-5pt\raisebox{-32.40352pt}{\includegraphics{figures_tikz/fig_ablation_deepblending.pdf}}\kern-5pt}\ht0=\dimexpr\ht0-5pt\relax\dp0=\dimexpr\dp0-5pt\relax\box0\prevdepth=27.40352pt\relax\fi%
\caption{Ablation of blending. We compare our \textit{softEDT}-based weight computation on the \textit{LiveNVS} dataset with \textit{Deep Blending}~\cite{hedman2018deep} by replacing weight computation in our pipeline with their learned weight prediction network.}%
    \label{fig:ablation_deepblending}
\end{figure}

Furthermore, two-band blending provides robustness towards calibration errors.
\figref{fig:jittered} displays qualitative examples of how errors in pose calibration can lead to ghosting artifacts when blending naively.
In this figure, input poses have been artificially jittered to simulate calibration errors. 
Our method shows versatility to still produce high fidelity results. While
$\sigma=0.95$ performs best in our experiments without artificial noise, increasing $\sigma$ may facilitate our methods robustness.
\section{Accompanying Video}%
\label{sup:sec:video}
The accompanying video shows a fly-through of the \mbox{\textit{Argoverse 2}} scene {\scriptsize\texttt{0a132537-3aec-35bb-af13-7faa0811000d}}.
It aims to demonstrate our results in a moving scenario, with a focus on visual quality.
In contrast to our evaluation, we pass all seven cameras to our approach. 
Although the dataset contains timestamps and is passed to our method in a streaming setup, the video is captured offline, i.e., higher frame rate.
The mentioned temporal anti-aliasing similar to \citet{lewis2025nepo} is enabled to reduce flickering.
At all times, the capturing sensors are at least $0.5$ \si{\second} ahead of the re-rendered viewport.
Thus the video displays an interpolation scenario on an automotive scene, which was not shown in the manuscript before.
Throughout the video, the capture sensor poses are visualized as white camera frustums, while the vehicle's dimensions are indicated by a white wireframe outline.

The video begins with the camera following the original capture trajectory before performing a full rotation around the vehicle. 
The camera then moves upward and backward, significantly departing from the capture trajectory.
Next, the vehicle is shown from a third-person perspective before executing a lane change.
Subsequently, the camera moves back and away from the original sensor positions. 
A second lane change is then performed before the camera returns to the original capture trajectory. 
 
\begin{table}[t]
\renewcommand{\arraystretch}{1.0}
\setlength{\tabcolsep}{4pt}
\centering
\begin{tabular}{l|ccc}
\toprule
 & \multicolumn{3}{c}{LiveNVS} \\
\midrule
Approach & PSNR$\uparrow$ & LPIPS$\downarrow$ & Time\,{\tiny \raisebox{0.3ex}{[\si{\milli\second}]}}\\
\midrule
Ours + \textit{Deep Blending}~\cite{hedman2018deep} & 21.84 & 0.2128 & 147 \\
Ours & \cellcolor{secondgreen}22.03 & \cellcolor{secondgreen}0.1987 & \cellcolor{secondgreen}12.11\\
\bottomrule
\end{tabular}
\caption{\label{tbl:ablation_deepblending}Quantitative results of our \emph{softEDT}-based blending compared to \textit{Deep Blending}'s~\cite{hedman2018deep} neural blending network.}
\end{table}

\end{document}